\pdfoutput=1

\documentclass[11pt]{article}

\usepackage[preprint]{acl}

\usepackage{times}
\usepackage{latexsym}

\usepackage[T1]{fontenc}

\usepackage[utf8]{inputenc}

\usepackage{microtype}

\usepackage{inconsolata}

\usepackage{graphicx}

\usepackage{todonotes}
\usepackage{adjustbox}
\usepackage{booktabs}
\usepackage{xcolor}
\usepackage{amssymb}
\usepackage{float}

\title{Persona-driven Simulation of Voting Behavior in the European Parliament with Large Language Models}

\author{
 \textbf{Maximilian Kreutner\textsuperscript{1}},
 \textbf{Marlene Lutz\textsuperscript{1}},
 \textbf{Markus Strohmaier\textsuperscript{1,2,3}}
\\
 \textsuperscript{1}University of Mannheim
 \textsuperscript{2}GESIS - Leibnitz Institute for the Social Sciences, Cologne
 \\
 \textsuperscript{3}CSH Vienna
\\
 \small{
    \textbf{Correspondence:}
    \href{mailto:maximilian.kreutner@uni-mannheim.de}{maximilian.kreutner@uni-mannheim.de}
   }
 }

\begin{document}
\maketitle
\begin{abstract}
Large Language Models (LLMs) display remarkable capabilities to understand or even produce political discourse but have been found to consistently exhibit a progressive left-leaning bias. At the same time, so-called persona or identity prompts have been shown to produce LLM behavior that aligns with socioeconomic groups with which the base model is not aligned. In this work, we analyze whether zero-shot persona prompting with limited information can accurately predict individual voting decisions and, by aggregation, accurately predict the positions of European groups on a diverse set of policies.
We evaluate whether predictions are stable in response to counterfactual arguments, different persona prompts, and generation methods. Finally, we find that we can simulate the voting behavior of Members of the European Parliament reasonably well, achieving a weighted F1 score of approximately 0.793. Our persona dataset of politicians in the 2024 European Parliament and our code are available at the following url: \url{https://github.com/dess-mannheim/european_parliament_simulation}.

\end{abstract}

\section{Introduction}

\emph{Large Language Models (LLMs)} are trained on vast amounts of diverse data, encompassing a wide range of attitudes, opinions, and values. Given their broad exposure, it remains unclear to what extent these models inherit various perspectives and how these perspectives might resurface in their responses. As a result, there is a growing body of research dedicated to investigating the latent viewpoints of LLMs, particularly in relation to their political leaning \cite{feng2023pretraining, wright-etal-2024-llm, rottger2024political}. Studies have found that, by default, LLMs tend to exhibit progressive, left-center political positions \cite{rutinowski2024selfperception, taubenfeld2024systematic, hartmann2023political, socsci12030148}. 
At the same time, researchers have demonstrated that when properly adapted (e.g., by fine-tuning or prompting), LLMs can effectively understand \cite{bernardelle2025mapping, heseltine2024large, wu2023large} and reproduce \cite{bachmann2025adaptive, chalkidis2024llamaEU} the positions of various political groups. 

\textbf{Research objective:} Building on these early demonstrations, our work systematically explores the extent to which LLMs can represent the political votes of \textit{individual} politicians. Specifically, we aim to simulate the voting behavior of all \emph{Members of the European Parliament (MEPs)} in 2024 by leveraging the ability of LLMs to role-play as different personas. This type of persona prompting has been shown to allow LLMs to make predictions that better align with the points of view of various sociodemographic groups in certain contexts \cite{argyle2023onemany, tseng2024twopersona}.

The case of the European Parliament is particularly interesting, as MEPs are often part of both a national party and a European group. Therefore, MEPs' personal preferences and national party lines can significantly influence the decision to dissent from their corresponding European Group \cite{willumsen2023policy}. Although individual voting decisions in the early 2000s were still mostly determined by the party lines of national parties \cite{hix2002parliamentary}, the European Parliament has since undergone a phase of "Empowerment," where majority coalitions are policy-specific, which is unique compared to most national parliaments \cite{hix2013empowerment}.

\begin{figure*}[t]
    \centering
    \includegraphics[width=\linewidth]{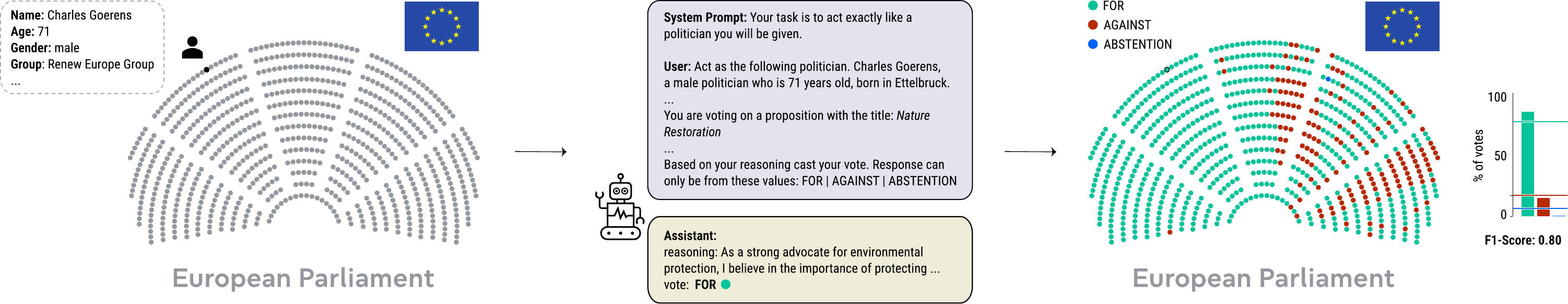}
    \caption{\textbf{Simulating the EU parliament.} We prompt the LLM to adopt the identity of each member of the European parliament. After providing the LLM with information about a proposal, we ask it to cast a vote in favor (FOR) or against (AGAINST) a proposal, or to abstain (ABSTENTION) from voting. We find that by using persona prompting, we can approximate the individual voting behavior of the members of the EU parliament, achieving a weighted F1 score of 0.80.}
    \label{fig:first_figure}
\end{figure*}

\textbf{Approach:} We use personas to prompt four LLMs to adopt the identity of each MEP and investigate the extent to which these models can simulate their voting behavior (cf. Figure \ref{fig:first_figure}). This includes predicting the individual decision-making of MEPs, who often balance competing interests with the group line on a given policy. We compare different models, persona prompts, and generation methods. We perform additional ablation studies to determine which attributes are the best predictors for LLMs.

\textbf{Results:} We find that in the majority of cases, persona-driven LLMs can accurately predict individual votes, achieving a weighted F1 score of 0.793; however, they do not predict abstentions well.  Additionally, persona-driven LLMs are able to correctly predict group positions on a diverse set of topics while also diverging from them at a rate comparable to real-world politicians. Our ablation study supports current political research, as the national party is the strongest predictor of the final vote.

In persona-driven simulations, we find evidence supporting three key observations. First, more detailed personas can help weaker models generate more accurate predictions. Second, prompting models with open-text reasoning before forcing multiple-choice options steers models more effectively. Third, models struggle to convincingly replicate the views of personas that diverge from their inherent biases.

\section{Related Work}

Predicting politicians' voting behavior is an active research area in both political science and NLP. Early work used word frequency, TF-IDF, fine-tuned embeddings, CNNs, or LSTMs to analyze bills and politician profiles \cite{gerrish2011predicting, kraft2016embedding, nay2017predicting, bonica2018inferring, kornilova2018party, yang2021joint}. Recent methods incorporate social media and socioeconomic data with graph-based models for richer context \cite{karimi2020multi, mou2021align, feng2022par, mou2023uppam}. Agent Based approaches with Large Language Models were first utilized by \citet{li2025political} and have shown to outperform previous methods. However, existing studies focus solely on the simpler two-party U.S. system. For the European Parliament, only traditional Machine Learning methods, e.g., Random Forest, have been tested and have been shown to perform only slightly better than the majority baseline \cite{guadarrama2025knowing}.
In addition, most recent approaches depend on access to a vast amount of past voting information or social media data for each politician, which is not always feasible in real-world scenarios. In contrast, our approach requires only publicly available information about politicians and any form of debate content.

Previous research has consistently shown that various LLMs have a bias toward progressive left-of-center political views by evaluating them on survey questions or different downstream tasks \cite{chalkidis2024llamaEU, chalkidis2024investigating, taubenfeld2024systematic, rutinowski2024selfperception, batzner2024germanpartiesqa,  feng2023pretraining,  hartmann2023political}. Furthermore, human feedback-tuned LLMs show an even larger bias compared to base models \cite{potter2024hidden, santurkar2023opinions}.



Several studies have demonstrated that LLMs can engage in "role-playing" by adopting specific personas within their contextual framework \cite{tseng2024twopersona, chen2024persona, hu2024quantifying, beck2024sensitivity}. In fact, they are so proficient at this that responses generated while impersonating politicians and other public figures are often perceived as more authentic than those given by the individuals themselves \cite{herbold2024large}. This role-playing capability has numerous applications. For instance, by using "silicon samples" and personas modeled with various socioeconomic attributes, LLMs can effectively replicate the viewpoints of demographically diverse subpopulations or predict election and voting behavior \cite{ma2024algorithmic, heyde2024voxpopuli, argyle2023onemany, yu2024large, zhang2024electionsim, bradshaw2024llm, yang2024llm, jiang2024donald}. Additionally, generating agents based on long form interviews enables LLMs to reproduce participants' responses on the General Social Survey with $85\%$ accuracy, compared to their own answers given two weeks later \cite{park2024generative}. We aim to use this capability to predict voting behavior across European groups, particularly those with whom LLMs are less aligned.

\section{Data Collection}

We collect vote information from HowTheyVote\footnote{\href{https://howtheyvote.eu/}{https://howtheyvote.eu/}}, a project dedicated to making European Parliament results transparently accessible to the public. Our focus is on \emph{roll-call votes}, the only type of vote in the European Parliament where the individual votes of all MEPs are recorded and published. We identify and select the most significant decisions of the 2024 parliament by checking if there is a press release and a debate about the vote. Our final dataset captures 27,770 individual voting decisions from 710 politicians across 47 key votes, which are displayed in Table \ref{tab:proposals_by_committee}. Three votes are resolutions initiated by the European Parliament itself, instead of proposals by the Committee. For the sake of simplicity, from this point on, we will still refer to all of these votes as \emph{proposals}.
Each MEP can vote in favor of (FOR), against (AGAINST), abstain (ABSTENTION) from a proposal, or not vote at all. While active abstentions are often strategic decisions \cite{muhlbock2017legislators}, the reasons for not voting are not recorded and can vary, such as illnesses or other responsibilities. As a result, we focus solely on predicting votes categorized as FOR, AGAINST, or ABSTENTION.

For all MEPs who voted on the 47 collected proposals, we retrieve their entries from English Wikipedia\footnote{\href{https://www.wikipedia.org/}{https://www.wikipedia.org/}} and Wikidata\footnote{\href{https://www.wikidata.org/}{https://www.wikidata.org/}}.
From the official website of the European Parliament\footnote{\href{https://www.europarl.europa.eu/meps/en/full-list/all}{https://www.europarl.europa.eu/meps/en/full-list/all}}, we further collect MEPs' affiliations with national parties, as well as all debates related to the selected proposals. 
We identify changes in party affiliation and membership in the European group using the HowTheyVote dataset and manually collect data for MEPs who have switched parties or groups.

\section{Methodology}
We explain our choice of models and how we included personas and proposal information.

\subsection{Models}
To prevent data leakage, we choose a model with a training cut-off prior to the earliest vote in our dataset on 16.01.2024: \emph{Llama3} \cite{dubey2024llama}. In addition, we chose a second model that was not developed in the west with comparable performance on the Open LLM Leaderboard \cite{openllmleaderboardv2}: \emph{Qwen2.5}\footnote{Qwen2.5's pretraining cutoff date is not officially disclosed. We considered the risk of data leakage but found it unlikely to be a confounding factor, as the results are comparably worse than LLama3.}  \cite{yang2024qwen2}. More specifically, we use the models:

- Llama3-70B (\texttt{Llama-3.1-70B-Instruct}),

- Llama3-8B (\texttt{Llama-3.1-8B-Instruct}),

- Qwen-72B (\texttt{Qwen2.5-72B-Instruct}),

- Qwen-7B (\texttt{Qwen2.5-7B-Instruct}). 

Additionally, we experimented with Mixtral \cite{jiang2024mixtral} as a European model but observed that it did not adhere well to our prompt format, as described in Appendix \ref{sec:other_models}. Therefore, we focus on LLama and Qwen for the main section of our paper. 

\subsection{Prompt Construction} \label{sec:prompt_construction}
To predict voting behavior, the LLM needs information about the identity of the MEP, the proposal on which they are expected to vote, and instructions on how to generate a response.

\textbf{Persona Construction.}
To create persona descriptions for each MEP, we consider two strategies. First, we create a combination of demographic and political affiliation attributes, including the full name, gender, age at the time of voting, birthplace, the country they represent, the European group, and the national party they are affiliated with at the time of voting. This approach allows us to precisely control the persona information that the LLM is provided with and enables us to estimate the influence of individual attributes on voting behavior.
Second, we use Llama3-70B to summarize the English Wikipedia article associated with each MEP. We investigate whether providing the LLM with longer contextual persona information can improve voting accuracy, similar to the approach used by \citet{park2024generative}, who used long form interview transcripts to simulate personas.

\textbf{Proposal Information}.
It is challenging to provide a short and unbiased description of proposals. The name of the proposal often lacks sufficient detail; for example, one proposal in our dataset is simply titled "Driving licenses". We considered using the official document for each proposal but found that the consolidated legislative document is too long for the context window of LLMs. On the other hand, official press releases, news reports, and summaries provided by the European Committee typically focus only on the positive aspects of the proposals, which could potentially bias the LLMs to always vote in favor of the proposals.
Instead, we adopt the theory that a common group position is established by the group's policy experts and then later adopted by other MEPs \cite{ringe2009decides}. We choose to use the speeches of each group representative, who are allocated speaking time before each vote to present the group's position on the respective proposal. The speeches are presented to the LLM in a randomized order, without disclosing the speaker's identity or group affiliation. Names of politicians, GROUP, and parties appearing in speeches were anonymized by substituting them with placeholders (e.g., [GROUP]; cf. Appendix \ref{sec:attribute_matching}).
We aim to provide a balanced overview of the pro- and contra-arguments regarding the implementation of the proposal while avoiding directly revealing each group's position. We note that, on average, each speech is nearly 1,500 characters long and constitutes the majority of the prompt.

\textbf{Response Format}.
We instruct the model to respond with exactly one of the options: FOR, AGAINST, or ABSTENTION. We further compare two different prompting strategies; \emph{reasoning (r)}, where the model responds first with an open text reasoning chain before choosing one of the options, and \emph{no reasoning (nr)}, where the model answers with one of the options directly. We prompt the model three times for each persona. In all runs, we use a temperature of $0.6$. Our full prompts can be found in Appendix \ref{sec:prompting_examples}.

\section{Results}
Since the distribution of ground truth votes is heavily imbalanced, with approximately $~77\%$ FOR votes, $~17\%$ AGAINST votes, and $~6\%$ ABSTENTION votes, we decide on the mean weighted F1 score across three runs as our evaluation measure. We report that the responses of all models are robust; i.e., in $87.1\%$ of cases, the models predict the same vote for each persona across all three runs, leading to a low standard deviation of $<0.002$ across all runs.
We analyze performance across three categories:

\begin{figure*}[t]
    \centering
    \includegraphics[width=\linewidth]{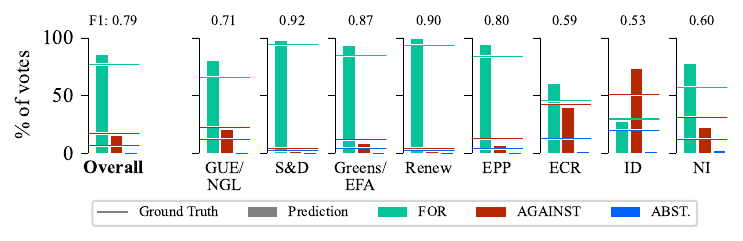}
    \caption{\textbf{Distribution of predicted votes per European group.} We display the voting predictions of the best performing model (Llama3-70B with attribute prompting and reasoning) compared to the ground truth. The weighted F1 score is displayed above each group. The model predicts the votes of center-left and progressive groups (S\&D, Renew, Greens/EFA) the best and performs worst for groups at the edge of the political spectrum (ID, GUE/NGL, ECR). Notably, the model almost never predicts abstentions.}
    \label{fig:best_results}
\end{figure*}

\emph{Model:} The prediction model, Llama3-8B, Llama3-70B, Qwen-7B or Qwen-72B.

\emph{Persona Prompt:} The persona description, either the Wikipedia summary or the attribute prompt.

\emph{Reasoning Strategy:} The reasoning strategy, either reasoning (r) or no reasoning (nr).

\begin{table}[!tbp]
\centering
\begin{adjustbox}{width={\columnwidth}}

\begin{tabular}{lcccc}
\toprule
Persona descr. & Llama3-70B & Llama3-8B & Qwen-72B & Qwen-7B \\
\midrule
all attributes (r) & \textbf{0.793} & 0.728 & 0.789 & 0.670 \\
wikipedia (r) & 0.779 & 0.724 & 0.773 & 0.697 \\
\midrule
all attributes (nr) & 0.772 & 0.687 & 0.773 & 0.688\\
wikipedia (nr) & 0.770 & 0.701 & 0.760 & 0.712 \\
\bottomrule
\end{tabular}
\end{adjustbox}
\caption{\textbf{Prediction performance of voting behavior with persona prompts.} We report the weighted F1 scores for voting prediction across all proposals for prompts with reasoning (r) and no reasoning (nr). While all predictions are better than the majority baseline (always predicting the majority vote leads to a weighted F1 score of 0.666), we find that a bigger model size in combination with reasoning produces the best simulation results.}
\label{tab:main_results}

\end{table}

\subsection{Persona Results}

We show our main results in Table \ref{tab:main_results}. The highest weighted F1-score of $0.793$ is achieved by Llama3-70B when employing reasoning with attribute-based persona prompts. Consistent with expectations, larger models surpass their smaller counterparts in performance. Furthermore, the western Llama models typically outperform Qwen models from China of comparable size.
The reasoning chain prompting approach improves overall prediction performance for LLama3-70B, Llama3-8b, and Qwen-72B. Notably, Qwen-7B performs worse overall with reasoning prompting. However, while the overall weighted F1-score does not improve with reasoning, it significantly alters prediction behavior. Specifically, without reasoning and the wikipedia prompts, Qwen-7B predicts FOR in $95.1\%$ of votes and correctly identifies only $19$ ABSTENTION instances, resulting in an F1-score of just $0.02$ for this class. Introducing reasoning reduces FOR predictions to $83.7\%$ of votes and, crucially, increases the correct identification of ABSTENTION instances to $199$, thereby improving the ABSTENTION F1-score to $0.110$. The AGAINST F1-score also shows an increase from $0.220$ to $0.280$. This demonstrates that, for Qwen-7B, the reasoning prompt induces a trade-off: while the aggregate score does not improve, it sacrifices performance on the majority class for a notable increase in sensitivity towards the minority classes.

Interestingly, the type of persona prompts impacts models in different ways. While the smaller models LLama3-8B and Qwen-7B can benefit from more detailed Wikipedia prompts, the larger models perform better with short attribute prompts.

We note that the best performing approach also correctly predicts the \emph{group line}, which we define as the majority position of a group per proposal, in $86.32\%$ of cases. When excluding ABSTENTION group lines, which were never predicted as the majority position of a group by the model, $90.16\%$ of the group lines are correctly predicted. The full group names and exact group line predictions compared to the actual group lines are shown in Appendix \ref{sec:groups}. The group lines of center-left and progressive parties are predicted correctly in most cases (Renew, S\&D, Greens/EFA), indicating that most prediction mistakes come from politicians who vote against the group line. On the other hand, for groups at the edge of the political spectrum (GUE/NGL, ECR, ID), the group line was often inaccurately predicted.

Next, we show the distribution of vote predictions disaggregated by European group in Figure \ref{fig:best_results}. We observe that the model consistently overestimates the number of FOR votes and significantly underestimates the number of ABSTENTION votes. We note that this trend holds across both Llama models. While the Qwen models predict more ABSTENTIONs, they have low precision, with the best model achieving $0.129$ and recall ($0.095$) when doing so (cf. Figure \ref{fig:best_results_qwen}). All models predict the votes of center-left and progressive groups (S\&D, Renew, Greens/EFA) the most accurately and perform worst for groups at the edge of the political spectrum (ID, ECR, GUE/NGL). This aligns with previous research on public opinion \cite{ma2024algorithmic}.

\subsection{Default Model Bias}

In addition to prompting LLMs to adopt the identity of a specific MEP, we also prompt LLMs directly with the default system prompt \textit{"You are a helpful assistant."} and reasoning chains. We find that both Llama3-70B and Qwen-72B vote in favor of all but four proposals, while Llama3-8B votes in favor of all but one proposal. This voting behavior aligns most with the group lines of S\&D and Renew. Qwen-7B votes in favor of 39 of 47 proposals, abstains in five, and votes against in three, which aligns closest to Greens/EFA. This behavior is in line with previous research \cite{rutinowski2024selfperception, santurkar2023opinions} that found a progressive and left-centered bias in many large language models.

\section{Ablation Studies}

We conduct two different ablation studies to examine whether the models are influenced more by the debates or the personas, and to identify which attributes most effectively improve performance. First, we remove all actual debates and replace them with counterfactual ones. Second, we remove specific attributes to determine their influence.

\subsection{Counterfactual Debates}

One issue that could cause the model to disproportionately predict votes in favor of a proposal may stem from the way the content of the proposals is introduced (cf. section \ref{sec:prompt_construction}). When prompting Llama3-70B to determine the stance of each speech used to inform about a proposal, the model classifies nearly $67\%$ of speeches as being in favor and only about $32\%$ as being against the respective proposal (roughly $1\%$ are classified as neutral). The fact that the majority of the speeches frame proposals positively could lead to the model being biased toward predicting votes in favor of a proposal. 

To control for this effect, we replace the actual debates with counterfactual ones. We instruct Llama3-70B to generate speeches that take the opposite stance of the original speeches and prompt it to use only the information provided in the original speech. To verify whether the generated counterfactual speeches accurately reflect the opposite stance, three voluntary annotators rate a randomly chosen speech for each of the 47 proposals. Based on a majority vote, 43 out of 47 speeches indeed align with the opposite stance relative to the original speeches, with a high level of agreement among annotators (Fleiss' Kappa score of $0.800$).

When prompting the models without personas, the number of votes in favor of proposals reduces from 43 to 32 for Llama3-70B and from 46 to 38 for Llama3-8B, showing that the debate context can have a limited influence on the final vote choice for these models.
On the other hand, voting behavior for both Qwen-72B and Qwen-7B does not change.

For computational reasons, we restrict our experiments with counterfactual speeches with personas to attribute prompting and Llama3-70B using the no reasoning approach.
We observe that using counterfactual speeches to describe the proposals decreases the prediction performance to a weighted F1 score of $0.753$. However, this still outperforms all Llama3-8B approaches with the original speeches (cf. Table \ref{tab:main_results}), indicating that the counterfactual debates still provide general information about the proposals.

\begin{figure}[!tbp]
    \centering
    \begin{adjustbox}{width={\columnwidth}}
    \includegraphics{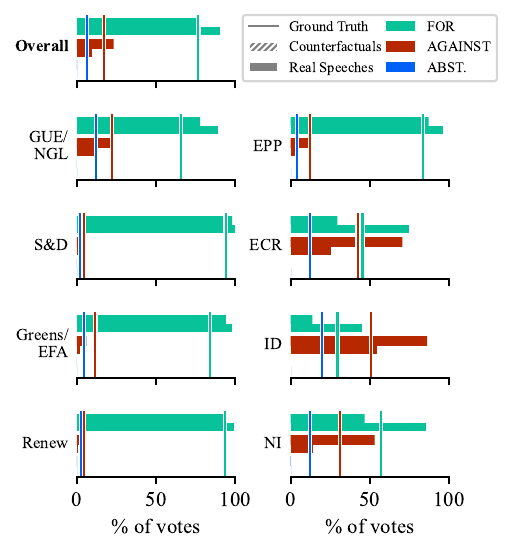} 
    \end{adjustbox}
    \caption{\textbf{Influence of counterfactual speeches on vote prediction across groups.} We compare the voting behavior of Llama3-70B when prompting with counterfactual speeches as opposed to the original speeches.  Personas based on MEPs affiliated with groups at the edges of the political spectrum (GUE/NGL, ID, ECR) tend to change their votes more frequently.}
    \label{fig:voting_counter_to_real_70}
\end{figure}

Figure \ref{fig:voting_counter_to_real_70} shows that prompting the model with counterfactual speeches indeed removes the FOR bias and overall slightly introduces an AGAINST bias. However, this effect is not uniform across groups. While far-left and far-right groups (GUE/NGL, ECR, ID) vote AGAINST significantly more often, the voting behavior of centrist groups (S\&D, Renew, Greens/EFA) remains largely unchanged. This suggests that personas continue to play a substantial role in the final vote prediction, despite the stance expressed in the speeches. Another group that votes AGAINST significantly more often is the group of Non-Inscrits (NI), who are not part of any major political group.

\subsection{Influence of different attributes}
\label{sec:attributes}

As we create personas for politicians of public interest, the models will likely have information about them in their training data. We are interested in whether the model can make good predictions even when only given the name of the politician in context. If additional attributes improve performance, we want to determine which ones enhance prediction capability the most. Therefore, for each MEP, we prompt the model with (i) only the name, (ii) the name and a single attribute, and (iii) the name and all attributes. To reduce computation costs, we restrict these simulations to Llama3-8B with reasoning, as Qwen-7B showed worse performance with attributes. 
Prompting with only the name of the MEP (weighted F1 score 0.681) already performs marginally better than the majority baseline (0.666). Prompting with all attributes yields the best performance (0.728). Incorporating only the name and the national party (0.722) or the group (0.718) into the persona prompt results in the most significant improvement compared to prompting with the name only, and it nearly matches the effectiveness of prompting with all attributes.

\begin{figure}[!tbp]
    \centering
    \begin{adjustbox}{width={\columnwidth}}
    \includegraphics{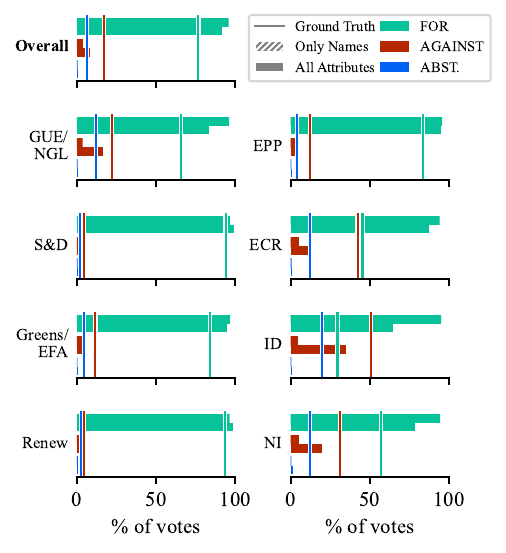} 
    \end{adjustbox}
    \caption{\textbf{Influence of the persona description on vote prediction across groups.} We illustrate the vote distribution when prompting Llama3-8B with only the name vs prompting with all persona attributes. Providing all attributes of a  persona significantly changes the voting behavior of the model compared to only providing the name.}
    \label{fig:voting_name_to_all}
\end{figure}

Figure \ref{fig:voting_name_to_all} shows that prompting with all attributes, compared to providing only the name, significantly steers the model towards a more accurate simulation for groups on the political edge, but also incorrectly steers personas of center left parties to vote in favor of even more proposals.

In addition to analyzing the votes cast by the LLM, we analyze the reasoning chains it provides to justify the voting decision. Specifically, we investigate which persona attributes are explicitly mentioned in the reasoning chains to estimate how much the LLM focuses on the characteristics of an individual persona (cf. Appendix \ref{sec:attribute_matching} for details). 
We emphasize that the reasoning chains should not be interpreted as the LLMs' internal decision making process, as they may significantly misrepresent the actual reasons behind the predictions of a model \cite{turpin2023dontsay}. However, we hypothesize that the presence of certain attributes in the reasoning chains may still influence the voting decisions, as transformer models generate content in an autoregressive manner.

As shown in Table \ref{tab:attribute_in_reasonings}, the attributes provided in the prompt also appear more often in the reasoning chains. When all attributes are given, the national party is mentioned most often, followed by the European group and the country. Considering this, along with the fact that using only the national party along with the name achieves almost the same performance as the entire set of attributes, this supports the argument of current political research, which indicates that the primary principal of MEPs is their national party \cite{willumsen2023policy}.

The results in Table \ref{tab:attribute_in_reasonings} for prompting with only the name further indicate that the model has knowledge about MEPs to some extent. For 150 of 653 MEPs in groups, the LLM mentioned the correct European group in the reasoning chains at least once. We assume that the model's pre-training data contains more information than the information we provide in the prompt. However, prediction results still improve when using all attributes for these politicians compared to using only the name, indicating that persona attributes have an additional steering effect, even when the model's pre-training data contains information about the person.

\begin{table}[!tbp]
\centering
\begin{adjustbox}{width={\columnwidth}}
\begin{tabular}{lrrrrrr}
\toprule
\textbf{Persona descr.} & \textbf{age} & \textbf{birthplace} & \textbf{country} & \textbf{gender} & \textbf{group} & \textbf{nat. party} \\
\midrule
name only & 0.07 & 0.06 & \textbf{3.29} & 2.74 & 0.73 & 0.14 \\
\midrule
age & \textbf{2.92} & 0.07 & 3.83 & 2.51 & 0.56 & 0.12 \\
birthplace & 0.03 & 20.30 & \textbf{23.55} & 2.61 & 0.95 & 0.14 \\
country & 0.08 & 0.53 & \textbf{60.88} & 2.74 & 0.15 & 0.04 \\
gender & 0.05 & 0.03 & 2.15 & \textbf{2.53} & 0.23 & 0.08 \\
group & 0.01 & 0.01 & 0.74 & 2.55 & \textbf{49.43} & 0.09 \\
nat. party & 0.01 & 0.06 & 7.22 & 2.63 & 1.12 & \textbf{55.41} \\
all attributes  & 0.02 & 0.11 & 16.67 & 2.57 & 31.08 & \textbf{51.42} \\

\bottomrule
\end{tabular}
\end{adjustbox}
\caption{\textbf{Attribute mentions in the reasoning chains for different persona descriptions.} We compute the percentage of reasoning chains of Llama3-8B that contain a persona attribute for different persona prompts. We find that attributes that appear in the persona description have a higher likelihood to appear in the reasoning chains of the model as well, although to different extent. When prompted with all attributes, the model most often mentions the national party of a member, while age and birthplace are rarely included.}
\label{tab:attribute_in_reasonings}
\end{table}

In addition, we evaluate whether the model takes decisions solely based on either the national party or the group. For this, we calculate the median variance of both predicted and actual votes per group and national party across all proposals to check if the model's predictions are always the same for personas of the same national party or European group. The median variance is $0.11$ for national parties and $0.08$ for European groups. These values are only slightly lower than the median variance of actual votes, which are $0.13$ and $0.10$, respectively. The model simulates slightly more party cohesion than in reality, but not full cohesion.

\section{Discussion}

We see that all models struggle to predict ABSTENTION votes correctly. We hypothesize that this could be due to the following three reasons. First, only around $1\%$ of speeches that we leverage as descriptors for the different proposals are classified as neutral by Llama3-70B. We suggest that incorporating more speeches with a neutral stance (i.e., neither advocating FOR nor AGAINST a proposal) could increase the number of ABSTENTION votes. 

Second, our simulation setup does not take the strategic aspects of coalitions into account. We find that the majority of actual abstention votes are cast by MEPs from groups that form the opposition on most proposals. Groups that form coalitions cannot allow abstentions if they want to retain legislative power, whereas opposition groups have more leeway \cite{willumsen2012strategic}. 

Third, unlike real politicians, our models face no consequences for voting against the group line. Real politicians often have competing interests, which can explain why they may abstain from voting. For example, when the position of the European group conflicts with the national party, politicians are caught in a dilemma. They must balance loyalty to the European group, important for career advancement and influence, with loyalty to the national party, which controls the selection of candidates for future elections \cite{muhlbock2017legislators}. We assume that LLMs do not capture these complex dynamics and, as a result, are unlikely to take them into account when making predictions.

In addition to determining the overall predictive performance of LLMs, we find interesting takeaways for further persona based simulations. \citet{rottger2024political} have shown that models produce different answers on the Politicial Compass Test, when they respond in an open format before being forced to choose a multiple choice option. Conversely, in our approach, we have ground-truth data and can see that this open reasoning format improves performance across almost all models tested, especially for personas that do not have the same voting behavior as the base model. For Qwen-7B, it steers the model more, leading to better ABSTENTION and AGAINST performance. We argue that this should be kept in mind for further experiments.

When exposed to counterfactual debates, LLMs with \emph{politically divergent personas} (those whose voting behavior differs most from the model's default system prompt voting behavior, e.g., politicians from ECR or ID) change their predicted voting behavior much more significantly than LLMs with \emph{politically aligned personas} (those whose voting behavior matches the base model). Similar to findings in previous work \cite{taubenfeld2024systematic, liu2024evaluating}, this points towards an inability of LLMs to consistently replicate the viewpoints of politically divergent personas, as their fundamental position on a topic can be changed by counterfactual arguments, whereas LLMs prompted with politically aligned personas do not change their fundamental position as easily. However, stylistic differences beyond stance, such as speech length, can be confounders in our setup, and we call for further experiments in a controlled setup.

\section{Conclusion}

We simulate the individual voting behavior of MEPs by prompting LLMs with personas and set a baseline for vote prediction of politicians in the European Parliament. Our results align with current political research, which suggests that the national party is the primary principal for MEPs \cite{willumsen2023policy}. We find further evidence of a center-left leaning bias of LLMs, both in vote prediction and in staying true to the persona's position when presented with counterfactual debates. We find that persona-driven LLMs are more steerable when responding in open text format before answering a multiple choice question. This suggests that future work using personas should prioritize open-ended reasoning, particularly when employing smaller models. Additionally, we find that for public figures like politicians, who are likely part of the training data of LLMs, prompting with additional information can help steer the model more effectively. To support further research in political simulation, we release our personas and simulation code. We hope the personas can facilitate further research in other directions, and because our approach works with easily obtainable attribute data and general debates, it can be effective for other political systems as well.

\section*{Limitations}

\paragraph{Bias of roll-call votes:} As our computational capacity is limited, we had to select a subset of all roll-call votes. Therefore, we determined that roll-call votes that had their own official press release and a debate solely focused on their topic would be the most relevant and important votes. This, however, may miss certain important votes, e.g., a number of proposals that were introduced to reduce and monitor migration and asylum in the EU. None of these roll-call votes had their own press release or a standalone debate. Instead, there was one joint debate for the entire Migration and Asylum package. Therefore, these roll-call votes are not part of our experiments.

\paragraph{Machine translation:} As there is a multitude of languages spoken in the European Parliament and the majority of these languages are not supported by current LLMs, we use machine translation via google translate for the speeches. This can introduce bias into our debates and inaccurately reflect the viewpoints of certain countries. However, we argue that excluding speeches in low-resource languages would introduce an even greater bias.

\paragraph{Data Cutoff:} As we do not have access to the training data of the models, we cannot know if the voting results appear in the pre-training data of the models. We rely on information publicly provided about the data cutoff date by the creators, which we cannot verify ourselves.

\paragraph{Similarity of proposals:} Additionally, while these specific votes are after the pre-training cutoff date, there can be votes about similar proposals that can be part of the pre-training data.

\section*{Ethical Considerations}

We are aware that using LLMs to simulate democratic processes raises ethical questions in itself. Our primary goal is to see how well personas can eliminate the political bias inherent in LLMs and steer LLMs towards better predictions. We do not propose to use LLMs for actual political decision making. Here, we focus on the ethical problems that we encountered during our project, which we find particularly problematic. 
First, persona prompted LLMs can be used to impersonate real people and spread false information about politicians' motives to vote a certain way. Especially with reasoning prompts, we see arguments that contain information directly from the politician's life and sound plausible, but argue for a position the politician did not take in the vote.
Second, simply prompting an aligned LLM to create a counterfactual speech arguing against a real political speech can create harmful content; for example, giving an LLM the task of creating a speech for a politician arguing for banning forced labor products can result in a speech promoting forced labor as the economically responsible option. Additionally, these impersonated speeches generated by LLMs, based on real politicians, could be used to hijack democratic processes and spread false information. Debates could be reformulated by LLMs into speeches that are expected from a certain party or politician, which present different points and steer the discussion away from real arguments.

\bibliography{custom}

\appendix

\section{Groups of the European Parliament}
\label{sec:groups}

The European Parliament during our voting period consists of seven political groups and a set of Non-Inscrits (NI), who are not part of a major political group. The groups, ordered from political left-leaning to political right leaning, are: The Left group in the European Parliament (GUE/NGL), the Group of the Progressive Alliance of Socialists and Democrats in the European Parliament (S\&D), the Group of the Greens/European Free Alliance (Greens/EFA), the Renew Europe Group (Renew), the Group of the European People’s Party (EPP), the European Conservatives and Reformists Group (ECR), and the Identity and Democracy Group (ID). Note that this left to right ordering does not strictly hold true for all issues. We display exact group lines for each group and vote in our simulation in Table \ref{tab:comparison}.

\section{Performance per Group}

We show the prediction performance of our simulation for each group when simulating with Llama3-70B and Qwen-72B in Table \ref{tab:group_results}. Across all methods, the votes of center-left and progressive parties are predicted best (S\&D, Renew), while the votes of groups on the political edge (GUE/NGL, ECR, ID) are predicted worst. We see that personas can influence the voting behavior of LLMs, but similar to previous research, they cannot achieve high prediction performance for European groups with which the base model is not well aligned \cite{chalkidis2024investigating}.


\begin{table*}[tb]
\centering
\begin{adjustbox}{width=\linewidth}

\begin{tabular}{lccccccccc}
\toprule
Llama3-70B & \textbf{Overall} & GUE/NGL & S\&D & Greens/EFA & Renew & EPP & ECR & ID & NI \\

\midrule
all attributes (r) & \textbf{0.793} & 0.711 & 0.924 & 0.867 & 0.900 & 0.804 & 0.592 & 0.529 & 0.596 \\
wikipedia (r) & 0.779 & 0.669 & 0.919 & 0.849 & 0.894 & 0.795 & 0.543 & 0.538 & 0.602 \\
\midrule
all attributes (nr) & 0.772 & 0.671 & 0.911 & 0.804 & 0.899 & 0.792 & 0.529 & 0.560 & 0.569 \\
wikipedia (nr) & 0.770 & 0.662 & 0.924 & 0.854 & 0.895 & 0.786 & 0.507 & 0.528 & 0.588 \\
\bottomrule
\toprule
Qwen-72B & \textbf{Overall} & GUE/NGL & S\&D & Greens/EFA & Renew & EPP & ECR & ID & NI \\
\midrule
all attributes (r) & \textbf{0.789} & 0.735 & 0.917 & 0.842 & 0.899 & 0.799 & 0.574 & 0.574 & 0.563\\
wikipedia (r) & 0.773 & 0.703 & 0.918 & 0.855 & 0.892 & 0.787 & 0.502 & 0.548 & 0.576 \\
\midrule
all attributes (nr) & 0.773 & 0.687 & 0.912 & 0.846 & 0.900 & 0.777 & 0.516 & 0.578 & 0.525 \\
wikipedia (nr) & 0.760 & 0.664 & 0.915 & 0.849 & 0.899 & 0.771 & 0.475 & 0.523 & 0.559 \\
\bottomrule
\toprule
bert baseline & 0.762 & 0.615 & 0.903 & 0.860 & 0.892 & 0.823 & 0.536 & 0.454 & 0.413 \\
zero-shot baseline & 0.755 & 0.625 & 0.885 & 0.820 & 0.873 & 0.772 & 0.535 & 0.568 & 0.413 \\
majority baseline & 0.666 & 0.521 & 0.909 & 0.768 & 0.901 & 0.762 & 0.283 & 0.135 & 0.413 \\
\bottomrule
\end{tabular}
\end{adjustbox}
\caption{\textbf{Prediction performance of voting behavior across groups.} Weighted F1 scores for voting prediction with Llama3-70B and Qwen-72B across all proposals fore each group for prompts with reasoning (r), without reasoning (nr). For both models the reasoning and attribute approach performs best. Across all methods the votes of center-left and progressive parties are predicted the best (S\&D, Renew), while votes of groups of the political edge (GUE/NGL, ECR, ID) are predicted worst, altough Qwen's best approach performs slightly better for very edges (GUE/NGL, ID).}
\label{tab:group_results}
\end{table*}

\begin{figure*}[t]
    \centering
    \includegraphics[width=\linewidth]{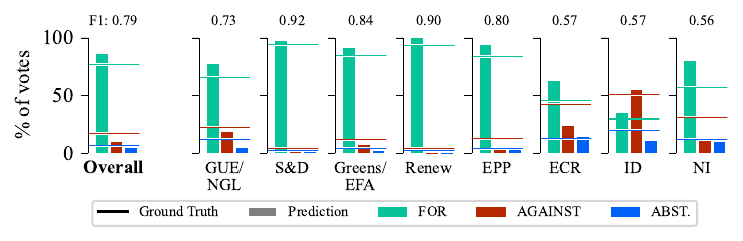}
    \caption{\textbf{Distribution of predicted votes per European group for Qwen-72B.} We display the voting predictions of the best performing Qwen approach (Qwen-72B with Attribute prompting and reasoning) compared to the ground truth. The weighted F1 score is displayed above each group. The model predicts the votes of center-left and progressive groups (S\&D, Renew, Greens/EFA) the best and performs worst for groups at the edge of the political spectrum (ID, GUE/NGL, ECR). Compared to LLama3-70B, the model does predict more ABSTENTIONS, however only seldom correctly (Precision 0.129 and Recall 0.095).}
    \label{fig:best_results_qwen}
\end{figure*}

Furthermore, we show the voting behavior of the best-performing Qwen approach with Wikipedia personas and reasoning in Figure \ref{fig:best_results_qwen}. Compared to the best Llama3-70B approach, the model predicts ABSTENTION more often but also struggles to predict them correctly, as both Precision (0.129) and Recall (0.095) are low for ABSTENTION votes.

\section{Stronger baselines}

To provide a comparison for our method, we established additional baselines using a simplified task. In this setup, we analyze the sentiment of the main spokesperson's speech for each group. If the spokesperson's speech is positive, we assign a \texttt{FOR} stance to their entire group. This baseline has an advantage because it relies on the spokesperson's group identity, which our main approach does not utilize. Since the NI group does not have a spokesperson, we simply predict \texttt{FOR} for the politicians of that group.

We use two approaches for this sentiment analysis:
\textbf{1. Fine-tuned BERT:} We fine-tune \texttt{bert-base-uncased} \cite{devlin2019bert} on 15,389 speeches from the years 2021 to 2023. To avoid data leakage, this training data is separate from our prediction dataset (2024). We use the MEP's voting decision as the ground truth label; for example, if an MEP voted \texttt{FOR}, we label their speech as positive. We train the model for 3 epochs with a batch size of 16 and a learning rate of $2e-5$.
\textbf{2. Zero-shot LLM:} We use the zero-shot sentiment predictions generated by Llama3-70B, following the same setup described in our counterfactual debate ablation.

The results are shown in the lower part of Table \ref{tab:group_results}. The fine-tuned BERT model achieves a mean F1-score of 0.762, and the zero-shot approach achieves 0.755. Both scores are lower than our best performing method (0.793), even though these baselines rely on the strong signal from the group spokesperson.

\section{Statistical Analysis}
To test whether the differences in prediction performance in Table \ref{tab:main_results} are statistically significant, we conduct an aligned rank transform (ART) ANOVA to examine the effects of reasoning, persona prompt (i.e., attribute and Wikipedia prompting), and their interaction on performance (i.e., F1-score) across the four models.
For the Qwen models, no effects were significant. In contrast, both Llama models showed a significant interaction between reasoning and prompt type (p < 0.05), while neither factor had a significant main effect.
This indicates that the benefit of reasoning differs between attribute and Wikipedia prompts. Specifically, reasoning provides a larger performance improvement for attribute prompting.

\section{Counterfactual Debates}

We also simulated counterfactual debates using Llama3-8B with reasoning. As shown in Table \ref{tab:counterfactual_8B}, similar to Llama3-70B, personas representing political parties at the extremes of the spectrum tend to change their votes more frequently than those of the center-left. However, unlike the Llama3-70B simulation, we observe that far-right party personas (ID, ECR) not only shift their votes but also change from AGAINST votes—cast in response to actual debates—to FOR votes when exposed to counterfactual debates. This is unexpected, as most debates shift from a positive to a negative stance. We hypothesize that LLMs struggle to role-play personas that do not align with the model’s inherent biases, making them more susceptible to persuasion when confronted with different arguments. In contrast, personas that reinforce the model’s base biases remain more resistant to such changes.

\begin{table}[ht!]
    \centering
    \begin{adjustbox}{width={\columnwidth}}
    \begin{tabular}{lcc}
    \toprule
    group & \multicolumn{1}{p{2.5cm}}{\centering Median AGAINST to FOR percentage} & \multicolumn{1}{p{2.5cm}}{\centering Median FOR to AGAINST percentage} \\
    \midrule
    GUE/NGL & 0.000 & 6.061 \\
    S\&D & 0.000 & 0.000 \\
    Greens/EFA & 0.000 & 1.538 \\
    Renew  & 0.000 & 2.247 \\
    EPP & 0.000 & 4.762 \\
    ECR & 5.085 & 10.638 \\
    ID  & 14.286 & 14.894 \\
    NI  & 4.348 & 12.121 \\
    \bottomrule
    \end{tabular}
    \end{adjustbox}
    \caption{\textbf{Vote changes across groups with LLama3-8B:} Median percentage of votes that change either from AGAINST for real speeches to FOR for counterfactual debates or the other way around. Same as Llama3-70B we see the most vote changes with parties at the end of the political spectrum. On the other hand, unlike the Llama3-70B model, votes of the far right parties also change from AGAINST to FOR, even though the majority of debates change to a negative stance.}
    \label{tab:counterfactual_8B}
\end{table}

\section{Proposals and Committees}
\label{sec:proposals}

We show the full list of names of the proposals and the committees that initiated them in Table \ref{tab:proposals_by_committee}. We selected these proposals because they had a significant societal impact, evidenced by an official EU press release focused on that proposal decision and by the fact that they were the sole focus of one or more debates in the European Parliament.
We analyzed the group line for each proposal and compared it with the predicted group line of our best performing approach Llama3-70B using reasoning and attribute prompts. We show the comparison in Table \ref{tab:comparison}. We see that the model never predicts the ABSTENTION group line. The Renew group was predicted best, followed by S\&D, Greens/EFA and EPP. For the far-left group GUE/NGL, the model predicts more FOR group lines on topics that include Ukraine and policies that introduce more digital processes, which involve more data collection of citizens (e.g., European Health Data Space or Instant Payments in Euro). For the ECR group, the model predicts more agreement with proposals that aim to preserve nature and reduce emissions (e.g., Nature Restoration, Methane emissions reduction in the energy sector). The ID group has the group line of ABSTENTION most often, which the model did not predict at all. Further false predictions are again related to ecological proposals; however, FOR this group, the model predicts more AGAINST group lines than FOR group lines (e.g., Ozone Depleting Substances, Empowering consumers for the green transition). The vote breakdown indicates that our current models struggle to predict when politicians from centrist parties vote against their group's established position. Conversely, for groups on the political fringes, the models sometimes fail to predict the group line itself.

\begin{table*}[p]
\begin{adjustbox}{width={\textwidth}}
\begin{tabular}{p{10.5cm}|ccccccc}
\toprule
group & GUE/NGL & S\&D & Greens/EFA & Renew & EPP & ECR & ID \\
proposal & {\small Actual/Pred.} & {\small Actual/Pred.} & {\small Actual/Pred.} & {\small Actual/Pred.} & {\small Actual/Pred.} & {\small Actual/Pred.} & {\small Actual/Pred.} \\
\midrule
Ambient air quality and cleaner air for Europe. Recast & \textcolor[HTML]{6ebd9b}{$\checkmark$} / \textcolor[HTML]{6ebd9b}{$\checkmark$} & \textcolor[HTML]{6ebd9b}{$\checkmark$} / \textcolor[HTML]{6ebd9b}{$\checkmark$} & \textcolor[HTML]{6ebd9b}{$\checkmark$} / \textcolor[HTML]{6ebd9b}{$\checkmark$} & \textcolor[HTML]{6ebd9b}{$\checkmark$} / \textcolor[HTML]{6ebd9b}{$\checkmark$} & \textcolor[HTML]{9a371e}{$\times$} / \textcolor[HTML]{6ebd9b}{$\checkmark$} & \textcolor[HTML]{9a371e}{$\times$} / \textcolor[HTML]{9a371e}{$\times$} & \textcolor[HTML]{9a371e}{$\times$} / \textcolor[HTML]{9a371e}{$\times$} \\
Artificial Intelligence Act & \textcolor[HTML]{9a371e}{$\times$} / \textcolor[HTML]{9a371e}{$\times$} & \textcolor[HTML]{6ebd9b}{$\checkmark$} / \textcolor[HTML]{6ebd9b}{$\checkmark$} & \textcolor[HTML]{6ebd9b}{$\checkmark$} / \textcolor[HTML]{6ebd9b}{$\checkmark$} & \textcolor[HTML]{6ebd9b}{$\checkmark$} / \textcolor[HTML]{6ebd9b}{$\checkmark$} & \textcolor[HTML]{6ebd9b}{$\checkmark$} / \textcolor[HTML]{6ebd9b}{$\checkmark$} & \textcolor[HTML]{6ebd9b}{$\checkmark$} / \textcolor[HTML]{6ebd9b}{$\checkmark$} & \textcolor[HTML]{6ebd9b}{$\checkmark$} / \textcolor[HTML]{9a371e}{$\times$} \\
Authorisation and supervision of medicinal products for human use and governing rules for the European Medicines Agency & \textcolor[HTML]{6ebd9b}{$\checkmark$} / \textcolor[HTML]{6ebd9b}{$\checkmark$} & \textcolor[HTML]{6ebd9b}{$\checkmark$} / \textcolor[HTML]{6ebd9b}{$\checkmark$} & \textcolor[HTML]{6ebd9b}{$\checkmark$} / \textcolor[HTML]{6ebd9b}{$\checkmark$} & \textcolor[HTML]{6ebd9b}{$\checkmark$} / \textcolor[HTML]{6ebd9b}{$\checkmark$} & \textcolor[HTML]{6ebd9b}{$\checkmark$} / \textcolor[HTML]{6ebd9b}{$\checkmark$} & \textcolor[HTML]{6ebd9b}{$\checkmark$} / \textcolor[HTML]{6ebd9b}{$\checkmark$} & \textcolor[HTML]{9a371e}{$\times$} / \textcolor[HTML]{9a371e}{$\times$} \\
Combating violence against women and domestic violence & \textcolor[HTML]{6ebd9b}{$\checkmark$} / \textcolor[HTML]{6ebd9b}{$\checkmark$} & \textcolor[HTML]{6ebd9b}{$\checkmark$} / \textcolor[HTML]{6ebd9b}{$\checkmark$} & \textcolor[HTML]{6ebd9b}{$\checkmark$} / \textcolor[HTML]{6ebd9b}{$\checkmark$} & \textcolor[HTML]{6ebd9b}{$\checkmark$} / \textcolor[HTML]{6ebd9b}{$\checkmark$} & \textcolor[HTML]{6ebd9b}{$\checkmark$} / \textcolor[HTML]{6ebd9b}{$\checkmark$} & \textcolor[HTML]{4663ec}{$\blacksquare$} / \textcolor[HTML]{6ebd9b}{$\checkmark$} & \textcolor[HTML]{6ebd9b}{$\checkmark$} / \textcolor[HTML]{9a371e}{$\times$} \\
Common rules promoting the repair of goods & \textcolor[HTML]{6ebd9b}{$\checkmark$} / \textcolor[HTML]{6ebd9b}{$\checkmark$} & \textcolor[HTML]{6ebd9b}{$\checkmark$} / \textcolor[HTML]{6ebd9b}{$\checkmark$} & \textcolor[HTML]{6ebd9b}{$\checkmark$} / \textcolor[HTML]{6ebd9b}{$\checkmark$} & \textcolor[HTML]{6ebd9b}{$\checkmark$} / \textcolor[HTML]{6ebd9b}{$\checkmark$} & \textcolor[HTML]{6ebd9b}{$\checkmark$} / \textcolor[HTML]{6ebd9b}{$\checkmark$} & \textcolor[HTML]{6ebd9b}{$\checkmark$} / \textcolor[HTML]{6ebd9b}{$\checkmark$} & \textcolor[HTML]{6ebd9b}{$\checkmark$} / \textcolor[HTML]{6ebd9b}{$\checkmark$} \\
Cyber Resilience Act & \textcolor[HTML]{4663ec}{$\blacksquare$} / \textcolor[HTML]{6ebd9b}{$\checkmark$} & \textcolor[HTML]{6ebd9b}{$\checkmark$} / \textcolor[HTML]{6ebd9b}{$\checkmark$} & \textcolor[HTML]{6ebd9b}{$\checkmark$} / \textcolor[HTML]{6ebd9b}{$\checkmark$} & \textcolor[HTML]{6ebd9b}{$\checkmark$} / \textcolor[HTML]{6ebd9b}{$\checkmark$} & \textcolor[HTML]{6ebd9b}{$\checkmark$} / \textcolor[HTML]{6ebd9b}{$\checkmark$} & \textcolor[HTML]{6ebd9b}{$\checkmark$} / \textcolor[HTML]{6ebd9b}{$\checkmark$} & \textcolor[HTML]{4663ec}{$\blacksquare$} / \textcolor[HTML]{6ebd9b}{$\checkmark$} \\
Data collection and sharing relating to short-term accommodation rental services & \textcolor[HTML]{6ebd9b}{$\checkmark$} / \textcolor[HTML]{6ebd9b}{$\checkmark$} & \textcolor[HTML]{6ebd9b}{$\checkmark$} / \textcolor[HTML]{6ebd9b}{$\checkmark$} & \textcolor[HTML]{6ebd9b}{$\checkmark$} / \textcolor[HTML]{6ebd9b}{$\checkmark$} & \textcolor[HTML]{6ebd9b}{$\checkmark$} / \textcolor[HTML]{6ebd9b}{$\checkmark$} & \textcolor[HTML]{6ebd9b}{$\checkmark$} / \textcolor[HTML]{6ebd9b}{$\checkmark$} & \textcolor[HTML]{6ebd9b}{$\checkmark$} / \textcolor[HTML]{6ebd9b}{$\checkmark$} & \textcolor[HTML]{6ebd9b}{$\checkmark$} / \textcolor[HTML]{6ebd9b}{$\checkmark$} \\
Deepening EU integration in view of future enlargement & \textcolor[HTML]{9a371e}{$\times$} / \textcolor[HTML]{6ebd9b}{$\checkmark$} & \textcolor[HTML]{6ebd9b}{$\checkmark$} / \textcolor[HTML]{6ebd9b}{$\checkmark$} & \textcolor[HTML]{6ebd9b}{$\checkmark$} / \textcolor[HTML]{6ebd9b}{$\checkmark$} & \textcolor[HTML]{6ebd9b}{$\checkmark$} / \textcolor[HTML]{6ebd9b}{$\checkmark$} & \textcolor[HTML]{6ebd9b}{$\checkmark$} / \textcolor[HTML]{6ebd9b}{$\checkmark$} & \textcolor[HTML]{9a371e}{$\times$} / \textcolor[HTML]{9a371e}{$\times$} & \textcolor[HTML]{9a371e}{$\times$} / \textcolor[HTML]{9a371e}{$\times$} \\
Definition of criminal offences and penalties for the violation of Union restrictive measures & \textcolor[HTML]{9a371e}{$\times$} / \textcolor[HTML]{6ebd9b}{$\checkmark$} & \textcolor[HTML]{6ebd9b}{$\checkmark$} / \textcolor[HTML]{6ebd9b}{$\checkmark$} & \textcolor[HTML]{6ebd9b}{$\checkmark$} / \textcolor[HTML]{6ebd9b}{$\checkmark$} & \textcolor[HTML]{6ebd9b}{$\checkmark$} / \textcolor[HTML]{6ebd9b}{$\checkmark$} & \textcolor[HTML]{6ebd9b}{$\checkmark$} / \textcolor[HTML]{6ebd9b}{$\checkmark$} & \textcolor[HTML]{6ebd9b}{$\checkmark$} / \textcolor[HTML]{6ebd9b}{$\checkmark$} & \textcolor[HTML]{4663ec}{$\blacksquare$} / \textcolor[HTML]{6ebd9b}{$\checkmark$} \\
Driving licences & \textcolor[HTML]{4663ec}{$\blacksquare$} / \textcolor[HTML]{6ebd9b}{$\checkmark$} & \textcolor[HTML]{6ebd9b}{$\checkmark$} / \textcolor[HTML]{6ebd9b}{$\checkmark$} & \textcolor[HTML]{6ebd9b}{$\checkmark$} / \textcolor[HTML]{6ebd9b}{$\checkmark$} & \textcolor[HTML]{6ebd9b}{$\checkmark$} / \textcolor[HTML]{6ebd9b}{$\checkmark$} & \textcolor[HTML]{9a371e}{$\times$} / \textcolor[HTML]{6ebd9b}{$\checkmark$} & \textcolor[HTML]{9a371e}{$\times$} / \textcolor[HTML]{9a371e}{$\times$} & \textcolor[HTML]{9a371e}{$\times$} / \textcolor[HTML]{9a371e}{$\times$} \\
Ecodesign for Sustainable Products Regulation & \textcolor[HTML]{6ebd9b}{$\checkmark$} / \textcolor[HTML]{6ebd9b}{$\checkmark$} & \textcolor[HTML]{6ebd9b}{$\checkmark$} / \textcolor[HTML]{6ebd9b}{$\checkmark$} & \textcolor[HTML]{6ebd9b}{$\checkmark$} / \textcolor[HTML]{6ebd9b}{$\checkmark$} & \textcolor[HTML]{6ebd9b}{$\checkmark$} / \textcolor[HTML]{6ebd9b}{$\checkmark$} & \textcolor[HTML]{6ebd9b}{$\checkmark$} / \textcolor[HTML]{6ebd9b}{$\checkmark$} & \textcolor[HTML]{9a371e}{$\times$} / \textcolor[HTML]{6ebd9b}{$\checkmark$} & \textcolor[HTML]{9a371e}{$\times$} / \textcolor[HTML]{9a371e}{$\times$} \\
Economic governance: requirements for budgetary frameworks of the Member States & \textcolor[HTML]{9a371e}{$\times$} / \textcolor[HTML]{9a371e}{$\times$} & \textcolor[HTML]{6ebd9b}{$\checkmark$} / \textcolor[HTML]{6ebd9b}{$\checkmark$} & \textcolor[HTML]{9a371e}{$\times$} / \textcolor[HTML]{9a371e}{$\times$} & \textcolor[HTML]{6ebd9b}{$\checkmark$} / \textcolor[HTML]{6ebd9b}{$\checkmark$} & \textcolor[HTML]{6ebd9b}{$\checkmark$} / \textcolor[HTML]{6ebd9b}{$\checkmark$} & \textcolor[HTML]{6ebd9b}{$\checkmark$} / \textcolor[HTML]{6ebd9b}{$\checkmark$} & \textcolor[HTML]{9a371e}{$\times$} / \textcolor[HTML]{9a371e}{$\times$} \\
Empowering consumers for the green transition & \textcolor[HTML]{6ebd9b}{$\checkmark$} / \textcolor[HTML]{6ebd9b}{$\checkmark$} & \textcolor[HTML]{6ebd9b}{$\checkmark$} / \textcolor[HTML]{6ebd9b}{$\checkmark$} & \textcolor[HTML]{6ebd9b}{$\checkmark$} / \textcolor[HTML]{6ebd9b}{$\checkmark$} & \textcolor[HTML]{6ebd9b}{$\checkmark$} / \textcolor[HTML]{6ebd9b}{$\checkmark$} & \textcolor[HTML]{6ebd9b}{$\checkmark$} / \textcolor[HTML]{6ebd9b}{$\checkmark$} & \textcolor[HTML]{6ebd9b}{$\checkmark$} / \textcolor[HTML]{6ebd9b}{$\checkmark$} & \textcolor[HTML]{6ebd9b}{$\checkmark$} / \textcolor[HTML]{9a371e}{$\times$} \\
Energy Charter Treaty: withdrawal of the Union & \textcolor[HTML]{6ebd9b}{$\checkmark$} / \textcolor[HTML]{6ebd9b}{$\checkmark$} & \textcolor[HTML]{6ebd9b}{$\checkmark$} / \textcolor[HTML]{6ebd9b}{$\checkmark$} & \textcolor[HTML]{6ebd9b}{$\checkmark$} / \textcolor[HTML]{6ebd9b}{$\checkmark$} & \textcolor[HTML]{6ebd9b}{$\checkmark$} / \textcolor[HTML]{6ebd9b}{$\checkmark$} & \textcolor[HTML]{6ebd9b}{$\checkmark$} / \textcolor[HTML]{6ebd9b}{$\checkmark$} & \textcolor[HTML]{6ebd9b}{$\checkmark$} / \textcolor[HTML]{6ebd9b}{$\checkmark$} & \textcolor[HTML]{4663ec}{$\blacksquare$} / \textcolor[HTML]{6ebd9b}{$\checkmark$} \\
Energy performance of buildings & \textcolor[HTML]{6ebd9b}{$\checkmark$} / \textcolor[HTML]{6ebd9b}{$\checkmark$} & \textcolor[HTML]{6ebd9b}{$\checkmark$} / \textcolor[HTML]{6ebd9b}{$\checkmark$} & \textcolor[HTML]{6ebd9b}{$\checkmark$} / \textcolor[HTML]{6ebd9b}{$\checkmark$} & \textcolor[HTML]{6ebd9b}{$\checkmark$} / \textcolor[HTML]{6ebd9b}{$\checkmark$} & \textcolor[HTML]{6ebd9b}{$\checkmark$} / \textcolor[HTML]{6ebd9b}{$\checkmark$} & \textcolor[HTML]{9a371e}{$\times$} / \textcolor[HTML]{6ebd9b}{$\checkmark$} & \textcolor[HTML]{9a371e}{$\times$} / \textcolor[HTML]{9a371e}{$\times$} \\
Establishing the Strategic Technologies for Europe Platform (‘STEP’) & \textcolor[HTML]{9a371e}{$\times$} / \textcolor[HTML]{9a371e}{$\times$} & \textcolor[HTML]{6ebd9b}{$\checkmark$} / \textcolor[HTML]{6ebd9b}{$\checkmark$} & \textcolor[HTML]{6ebd9b}{$\checkmark$} / \textcolor[HTML]{6ebd9b}{$\checkmark$} & \textcolor[HTML]{6ebd9b}{$\checkmark$} / \textcolor[HTML]{6ebd9b}{$\checkmark$} & \textcolor[HTML]{6ebd9b}{$\checkmark$} / \textcolor[HTML]{6ebd9b}{$\checkmark$} & \textcolor[HTML]{6ebd9b}{$\checkmark$} / \textcolor[HTML]{6ebd9b}{$\checkmark$} & \textcolor[HTML]{6ebd9b}{$\checkmark$} / \textcolor[HTML]{9a371e}{$\times$} \\
Establishing the Ukraine Facility & \textcolor[HTML]{9a371e}{$\times$} / \textcolor[HTML]{6ebd9b}{$\checkmark$} & \textcolor[HTML]{6ebd9b}{$\checkmark$} / \textcolor[HTML]{6ebd9b}{$\checkmark$} & \textcolor[HTML]{6ebd9b}{$\checkmark$} / \textcolor[HTML]{6ebd9b}{$\checkmark$} & \textcolor[HTML]{6ebd9b}{$\checkmark$} / \textcolor[HTML]{6ebd9b}{$\checkmark$} & \textcolor[HTML]{6ebd9b}{$\checkmark$} / \textcolor[HTML]{6ebd9b}{$\checkmark$} & \textcolor[HTML]{6ebd9b}{$\checkmark$} / \textcolor[HTML]{6ebd9b}{$\checkmark$} & \textcolor[HTML]{4663ec}{$\blacksquare$} / \textcolor[HTML]{9a371e}{$\times$} \\
European Digital Identity framework & \textcolor[HTML]{9a371e}{$\times$} / \textcolor[HTML]{9a371e}{$\times$} & \textcolor[HTML]{6ebd9b}{$\checkmark$} / \textcolor[HTML]{6ebd9b}{$\checkmark$} & \textcolor[HTML]{9a371e}{$\times$} / \textcolor[HTML]{9a371e}{$\times$} & \textcolor[HTML]{6ebd9b}{$\checkmark$} / \textcolor[HTML]{6ebd9b}{$\checkmark$} & \textcolor[HTML]{6ebd9b}{$\checkmark$} / \textcolor[HTML]{6ebd9b}{$\checkmark$} & \textcolor[HTML]{9a371e}{$\times$} / \textcolor[HTML]{9a371e}{$\times$} & \textcolor[HTML]{9a371e}{$\times$} / \textcolor[HTML]{9a371e}{$\times$} \\
European Health Data Space & \textcolor[HTML]{9a371e}{$\times$} / \textcolor[HTML]{6ebd9b}{$\checkmark$} & \textcolor[HTML]{6ebd9b}{$\checkmark$} / \textcolor[HTML]{6ebd9b}{$\checkmark$} & \textcolor[HTML]{9a371e}{$\times$} / \textcolor[HTML]{6ebd9b}{$\checkmark$} & \textcolor[HTML]{6ebd9b}{$\checkmark$} / \textcolor[HTML]{6ebd9b}{$\checkmark$} & \textcolor[HTML]{6ebd9b}{$\checkmark$} / \textcolor[HTML]{6ebd9b}{$\checkmark$} & \textcolor[HTML]{6ebd9b}{$\checkmark$} / \textcolor[HTML]{6ebd9b}{$\checkmark$} & \textcolor[HTML]{9a371e}{$\times$} / \textcolor[HTML]{9a371e}{$\times$} \\
Extending the list of EU crimes to hate speech and hate crime & \textcolor[HTML]{6ebd9b}{$\checkmark$} / \textcolor[HTML]{6ebd9b}{$\checkmark$} & \textcolor[HTML]{6ebd9b}{$\checkmark$} / \textcolor[HTML]{6ebd9b}{$\checkmark$} & \textcolor[HTML]{6ebd9b}{$\checkmark$} / \textcolor[HTML]{6ebd9b}{$\checkmark$} & \textcolor[HTML]{6ebd9b}{$\checkmark$} / \textcolor[HTML]{6ebd9b}{$\checkmark$} & \textcolor[HTML]{6ebd9b}{$\checkmark$} / \textcolor[HTML]{6ebd9b}{$\checkmark$} & \textcolor[HTML]{9a371e}{$\times$} / \textcolor[HTML]{9a371e}{$\times$} & \textcolor[HTML]{9a371e}{$\times$} / \textcolor[HTML]{9a371e}{$\times$} \\
Fluorinated gases regulation & \textcolor[HTML]{6ebd9b}{$\checkmark$} / \textcolor[HTML]{6ebd9b}{$\checkmark$} & \textcolor[HTML]{6ebd9b}{$\checkmark$} / \textcolor[HTML]{6ebd9b}{$\checkmark$} & \textcolor[HTML]{6ebd9b}{$\checkmark$} / \textcolor[HTML]{6ebd9b}{$\checkmark$} & \textcolor[HTML]{6ebd9b}{$\checkmark$} / \textcolor[HTML]{6ebd9b}{$\checkmark$} & \textcolor[HTML]{6ebd9b}{$\checkmark$} / \textcolor[HTML]{6ebd9b}{$\checkmark$} & \textcolor[HTML]{9a371e}{$\times$} / \textcolor[HTML]{6ebd9b}{$\checkmark$} & \textcolor[HTML]{9a371e}{$\times$} / \textcolor[HTML]{9a371e}{$\times$} \\
Framework of measures for strengthening Europe’s net-zero technology products manufacturing ecosystem (Net Zero Industry Act) & \textcolor[HTML]{9a371e}{$\times$} / \textcolor[HTML]{9a371e}{$\times$} & \textcolor[HTML]{6ebd9b}{$\checkmark$} / \textcolor[HTML]{6ebd9b}{$\checkmark$} & \textcolor[HTML]{9a371e}{$\times$} / \textcolor[HTML]{6ebd9b}{$\checkmark$} & \textcolor[HTML]{6ebd9b}{$\checkmark$} / \textcolor[HTML]{6ebd9b}{$\checkmark$} & \textcolor[HTML]{6ebd9b}{$\checkmark$} / \textcolor[HTML]{6ebd9b}{$\checkmark$} & \textcolor[HTML]{9a371e}{$\times$} / \textcolor[HTML]{9a371e}{$\times$} & \textcolor[HTML]{6ebd9b}{$\checkmark$} / \textcolor[HTML]{9a371e}{$\times$} \\
Geographical Indications for wine, spirit drinks and agricultural products & \textcolor[HTML]{6ebd9b}{$\checkmark$} / \textcolor[HTML]{6ebd9b}{$\checkmark$} & \textcolor[HTML]{6ebd9b}{$\checkmark$} / \textcolor[HTML]{6ebd9b}{$\checkmark$} & \textcolor[HTML]{4663ec}{$\blacksquare$} / \textcolor[HTML]{6ebd9b}{$\checkmark$} & \textcolor[HTML]{6ebd9b}{$\checkmark$} / \textcolor[HTML]{6ebd9b}{$\checkmark$} & \textcolor[HTML]{6ebd9b}{$\checkmark$} / \textcolor[HTML]{6ebd9b}{$\checkmark$} & \textcolor[HTML]{6ebd9b}{$\checkmark$} / \textcolor[HTML]{6ebd9b}{$\checkmark$} & \textcolor[HTML]{6ebd9b}{$\checkmark$} / \textcolor[HTML]{6ebd9b}{$\checkmark$} \\
Inclusion of the right to abortion in the EU Charter of Fundamental Rights & \textcolor[HTML]{6ebd9b}{$\checkmark$} / \textcolor[HTML]{6ebd9b}{$\checkmark$} & \textcolor[HTML]{6ebd9b}{$\checkmark$} / \textcolor[HTML]{6ebd9b}{$\checkmark$} & \textcolor[HTML]{6ebd9b}{$\checkmark$} / \textcolor[HTML]{6ebd9b}{$\checkmark$} & \textcolor[HTML]{6ebd9b}{$\checkmark$} / \textcolor[HTML]{6ebd9b}{$\checkmark$} & \textcolor[HTML]{9a371e}{$\times$} / \textcolor[HTML]{9a371e}{$\times$} & \textcolor[HTML]{9a371e}{$\times$} / \textcolor[HTML]{9a371e}{$\times$} & \textcolor[HTML]{9a371e}{$\times$} / \textcolor[HTML]{9a371e}{$\times$} \\
Instant payments in euro & \textcolor[HTML]{4663ec}{$\blacksquare$} / \textcolor[HTML]{6ebd9b}{$\checkmark$} & \textcolor[HTML]{6ebd9b}{$\checkmark$} / \textcolor[HTML]{6ebd9b}{$\checkmark$} & \textcolor[HTML]{6ebd9b}{$\checkmark$} / \textcolor[HTML]{6ebd9b}{$\checkmark$} & \textcolor[HTML]{6ebd9b}{$\checkmark$} / \textcolor[HTML]{6ebd9b}{$\checkmark$} & \textcolor[HTML]{6ebd9b}{$\checkmark$} / \textcolor[HTML]{6ebd9b}{$\checkmark$} & \textcolor[HTML]{6ebd9b}{$\checkmark$} / \textcolor[HTML]{6ebd9b}{$\checkmark$} & \textcolor[HTML]{6ebd9b}{$\checkmark$} / \textcolor[HTML]{6ebd9b}{$\checkmark$} \\
Limit values for lead and its inorganic compounds and diisocyanates & \textcolor[HTML]{6ebd9b}{$\checkmark$} / \textcolor[HTML]{6ebd9b}{$\checkmark$} & \textcolor[HTML]{6ebd9b}{$\checkmark$} / \textcolor[HTML]{6ebd9b}{$\checkmark$} & \textcolor[HTML]{6ebd9b}{$\checkmark$} / \textcolor[HTML]{6ebd9b}{$\checkmark$} & \textcolor[HTML]{6ebd9b}{$\checkmark$} / \textcolor[HTML]{6ebd9b}{$\checkmark$} & \textcolor[HTML]{6ebd9b}{$\checkmark$} / \textcolor[HTML]{6ebd9b}{$\checkmark$} & \textcolor[HTML]{6ebd9b}{$\checkmark$} / \textcolor[HTML]{6ebd9b}{$\checkmark$} & \textcolor[HTML]{4663ec}{$\blacksquare$} / \textcolor[HTML]{9a371e}{$\times$} \\
Methane emissions reduction in the energy sector & \textcolor[HTML]{6ebd9b}{$\checkmark$} / \textcolor[HTML]{6ebd9b}{$\checkmark$} & \textcolor[HTML]{6ebd9b}{$\checkmark$} / \textcolor[HTML]{6ebd9b}{$\checkmark$} & \textcolor[HTML]{6ebd9b}{$\checkmark$} / \textcolor[HTML]{6ebd9b}{$\checkmark$} & \textcolor[HTML]{6ebd9b}{$\checkmark$} / \textcolor[HTML]{6ebd9b}{$\checkmark$} & \textcolor[HTML]{6ebd9b}{$\checkmark$} / \textcolor[HTML]{6ebd9b}{$\checkmark$} & \textcolor[HTML]{6ebd9b}{$\checkmark$} / \textcolor[HTML]{9a371e}{$\times$} & \textcolor[HTML]{9a371e}{$\times$} / \textcolor[HTML]{9a371e}{$\times$} \\
Nature restoration & \textcolor[HTML]{6ebd9b}{$\checkmark$} / \textcolor[HTML]{6ebd9b}{$\checkmark$} & \textcolor[HTML]{6ebd9b}{$\checkmark$} / \textcolor[HTML]{6ebd9b}{$\checkmark$} & \textcolor[HTML]{6ebd9b}{$\checkmark$} / \textcolor[HTML]{6ebd9b}{$\checkmark$} & \textcolor[HTML]{6ebd9b}{$\checkmark$} / \textcolor[HTML]{6ebd9b}{$\checkmark$} & \textcolor[HTML]{9a371e}{$\times$} / \textcolor[HTML]{9a371e}{$\times$} & \textcolor[HTML]{9a371e}{$\times$} / \textcolor[HTML]{9a371e}{$\times$} & \textcolor[HTML]{9a371e}{$\times$} / \textcolor[HTML]{9a371e}{$\times$} \\
Ozone depleting substances & \textcolor[HTML]{6ebd9b}{$\checkmark$} / \textcolor[HTML]{6ebd9b}{$\checkmark$} & \textcolor[HTML]{6ebd9b}{$\checkmark$} / \textcolor[HTML]{6ebd9b}{$\checkmark$} & \textcolor[HTML]{6ebd9b}{$\checkmark$} / \textcolor[HTML]{6ebd9b}{$\checkmark$} & \textcolor[HTML]{6ebd9b}{$\checkmark$} / \textcolor[HTML]{6ebd9b}{$\checkmark$} & \textcolor[HTML]{6ebd9b}{$\checkmark$} / \textcolor[HTML]{6ebd9b}{$\checkmark$} & \textcolor[HTML]{6ebd9b}{$\checkmark$} / \textcolor[HTML]{6ebd9b}{$\checkmark$} & \textcolor[HTML]{6ebd9b}{$\checkmark$} / \textcolor[HTML]{9a371e}{$\times$} \\
Packaging and packaging waste & \textcolor[HTML]{6ebd9b}{$\checkmark$} / \textcolor[HTML]{6ebd9b}{$\checkmark$} & \textcolor[HTML]{6ebd9b}{$\checkmark$} / \textcolor[HTML]{6ebd9b}{$\checkmark$} & \textcolor[HTML]{6ebd9b}{$\checkmark$} / \textcolor[HTML]{6ebd9b}{$\checkmark$} & \textcolor[HTML]{6ebd9b}{$\checkmark$} / \textcolor[HTML]{6ebd9b}{$\checkmark$} & \textcolor[HTML]{6ebd9b}{$\checkmark$} / \textcolor[HTML]{9a371e}{$\times$} & \textcolor[HTML]{9a371e}{$\times$} / \textcolor[HTML]{9a371e}{$\times$} & \textcolor[HTML]{9a371e}{$\times$} / \textcolor[HTML]{9a371e}{$\times$} \\
Plants obtained by certain new genomic techniques and their food and feed & \textcolor[HTML]{9a371e}{$\times$} / \textcolor[HTML]{9a371e}{$\times$} & \textcolor[HTML]{9a371e}{$\times$} / \textcolor[HTML]{9a371e}{$\times$} & \textcolor[HTML]{9a371e}{$\times$} / \textcolor[HTML]{9a371e}{$\times$} & \textcolor[HTML]{6ebd9b}{$\checkmark$} / \textcolor[HTML]{6ebd9b}{$\checkmark$} & \textcolor[HTML]{6ebd9b}{$\checkmark$} / \textcolor[HTML]{6ebd9b}{$\checkmark$} & \textcolor[HTML]{9a371e}{$\times$} / \textcolor[HTML]{6ebd9b}{$\checkmark$} & \textcolor[HTML]{6ebd9b}{$\checkmark$} / \textcolor[HTML]{6ebd9b}{$\checkmark$} \\
Preventing and combating trafficking in human beings and protecting its victims & \textcolor[HTML]{6ebd9b}{$\checkmark$} / \textcolor[HTML]{6ebd9b}{$\checkmark$} & \textcolor[HTML]{6ebd9b}{$\checkmark$} / \textcolor[HTML]{6ebd9b}{$\checkmark$} & \textcolor[HTML]{6ebd9b}{$\checkmark$} / \textcolor[HTML]{6ebd9b}{$\checkmark$} & \textcolor[HTML]{6ebd9b}{$\checkmark$} / \textcolor[HTML]{6ebd9b}{$\checkmark$} & \textcolor[HTML]{6ebd9b}{$\checkmark$} / \textcolor[HTML]{6ebd9b}{$\checkmark$} & \textcolor[HTML]{6ebd9b}{$\checkmark$} / \textcolor[HTML]{6ebd9b}{$\checkmark$} & \textcolor[HTML]{6ebd9b}{$\checkmark$} / \textcolor[HTML]{6ebd9b}{$\checkmark$} \\
Prevention of the use of the financial system for the purposes of money laundering or terrorist financing & \textcolor[HTML]{6ebd9b}{$\checkmark$} / \textcolor[HTML]{6ebd9b}{$\checkmark$} & \textcolor[HTML]{6ebd9b}{$\checkmark$} / \textcolor[HTML]{6ebd9b}{$\checkmark$} & \textcolor[HTML]{6ebd9b}{$\checkmark$} / \textcolor[HTML]{6ebd9b}{$\checkmark$} & \textcolor[HTML]{6ebd9b}{$\checkmark$} / \textcolor[HTML]{6ebd9b}{$\checkmark$} & \textcolor[HTML]{6ebd9b}{$\checkmark$} / \textcolor[HTML]{6ebd9b}{$\checkmark$} & \textcolor[HTML]{9a371e}{$\times$} / \textcolor[HTML]{6ebd9b}{$\checkmark$} & \textcolor[HTML]{4663ec}{$\blacksquare$} / \textcolor[HTML]{9a371e}{$\times$} \\
Prohibiting products made with forced labour on the Union market & \textcolor[HTML]{6ebd9b}{$\checkmark$} / \textcolor[HTML]{6ebd9b}{$\checkmark$} & \textcolor[HTML]{6ebd9b}{$\checkmark$} / \textcolor[HTML]{6ebd9b}{$\checkmark$} & \textcolor[HTML]{6ebd9b}{$\checkmark$} / \textcolor[HTML]{6ebd9b}{$\checkmark$} & \textcolor[HTML]{6ebd9b}{$\checkmark$} / \textcolor[HTML]{6ebd9b}{$\checkmark$} & \textcolor[HTML]{6ebd9b}{$\checkmark$} / \textcolor[HTML]{6ebd9b}{$\checkmark$} & \textcolor[HTML]{6ebd9b}{$\checkmark$} / \textcolor[HTML]{6ebd9b}{$\checkmark$} & \textcolor[HTML]{4663ec}{$\blacksquare$} / \textcolor[HTML]{6ebd9b}{$\checkmark$} \\
Report on the Commission’s 2023 Rule of Law report & \textcolor[HTML]{6ebd9b}{$\checkmark$} / \textcolor[HTML]{6ebd9b}{$\checkmark$} & \textcolor[HTML]{6ebd9b}{$\checkmark$} / \textcolor[HTML]{6ebd9b}{$\checkmark$} & \textcolor[HTML]{6ebd9b}{$\checkmark$} / \textcolor[HTML]{6ebd9b}{$\checkmark$} & \textcolor[HTML]{6ebd9b}{$\checkmark$} / \textcolor[HTML]{6ebd9b}{$\checkmark$} & \textcolor[HTML]{6ebd9b}{$\checkmark$} / \textcolor[HTML]{6ebd9b}{$\checkmark$} & \textcolor[HTML]{9a371e}{$\times$} / \textcolor[HTML]{9a371e}{$\times$} & \textcolor[HTML]{9a371e}{$\times$} / \textcolor[HTML]{9a371e}{$\times$} \\
Resolution on ongoing hearings under Article 7(1) TEU regarding Hungary to strengthen the rule of law and its budgetary implications & \textcolor[HTML]{6ebd9b}{$\checkmark$} / \textcolor[HTML]{6ebd9b}{$\checkmark$} & \textcolor[HTML]{6ebd9b}{$\checkmark$} / \textcolor[HTML]{6ebd9b}{$\checkmark$} & \textcolor[HTML]{6ebd9b}{$\checkmark$} / \textcolor[HTML]{6ebd9b}{$\checkmark$} & \textcolor[HTML]{6ebd9b}{$\checkmark$} / \textcolor[HTML]{6ebd9b}{$\checkmark$} & \textcolor[HTML]{6ebd9b}{$\checkmark$} / \textcolor[HTML]{6ebd9b}{$\checkmark$} & \textcolor[HTML]{9a371e}{$\times$} / \textcolor[HTML]{9a371e}{$\times$} & \textcolor[HTML]{9a371e}{$\times$} / \textcolor[HTML]{9a371e}{$\times$} \\
Resolution on the situation in Hungary and frozen EU funds & \textcolor[HTML]{7e7e7e}{$NO$} / \textcolor[HTML]{6ebd9b}{$\checkmark$} & \textcolor[HTML]{6ebd9b}{$\checkmark$} / \textcolor[HTML]{6ebd9b}{$\checkmark$} & \textcolor[HTML]{6ebd9b}{$\checkmark$} / \textcolor[HTML]{6ebd9b}{$\checkmark$} & \textcolor[HTML]{6ebd9b}{$\checkmark$} / \textcolor[HTML]{6ebd9b}{$\checkmark$} & \textcolor[HTML]{6ebd9b}{$\checkmark$} / \textcolor[HTML]{6ebd9b}{$\checkmark$} & \textcolor[HTML]{9a371e}{$\times$} / \textcolor[HTML]{6ebd9b}{$\checkmark$} & \textcolor[HTML]{9a371e}{$\times$} / \textcolor[HTML]{9a371e}{$\times$} \\
Shipments of waste & \textcolor[HTML]{6ebd9b}{$\checkmark$} / \textcolor[HTML]{6ebd9b}{$\checkmark$} & \textcolor[HTML]{6ebd9b}{$\checkmark$} / \textcolor[HTML]{6ebd9b}{$\checkmark$} & \textcolor[HTML]{6ebd9b}{$\checkmark$} / \textcolor[HTML]{6ebd9b}{$\checkmark$} & \textcolor[HTML]{6ebd9b}{$\checkmark$} / \textcolor[HTML]{6ebd9b}{$\checkmark$} & \textcolor[HTML]{6ebd9b}{$\checkmark$} / \textcolor[HTML]{6ebd9b}{$\checkmark$} & \textcolor[HTML]{6ebd9b}{$\checkmark$} / \textcolor[HTML]{6ebd9b}{$\checkmark$} & \textcolor[HTML]{6ebd9b}{$\checkmark$} / \textcolor[HTML]{9a371e}{$\times$} \\
Single Permit Directive. Recast & \textcolor[HTML]{6ebd9b}{$\checkmark$} / \textcolor[HTML]{6ebd9b}{$\checkmark$} & \textcolor[HTML]{6ebd9b}{$\checkmark$} / \textcolor[HTML]{6ebd9b}{$\checkmark$} & \textcolor[HTML]{6ebd9b}{$\checkmark$} / \textcolor[HTML]{6ebd9b}{$\checkmark$} & \textcolor[HTML]{6ebd9b}{$\checkmark$} / \textcolor[HTML]{6ebd9b}{$\checkmark$} & \textcolor[HTML]{6ebd9b}{$\checkmark$} / \textcolor[HTML]{6ebd9b}{$\checkmark$} & \textcolor[HTML]{9a371e}{$\times$} / \textcolor[HTML]{9a371e}{$\times$} & \textcolor[HTML]{9a371e}{$\times$} / \textcolor[HTML]{9a371e}{$\times$} \\
Situation of fundamental rights in the European Union - annual report 2022 and 2023 & \textcolor[HTML]{6ebd9b}{$\checkmark$} / \textcolor[HTML]{6ebd9b}{$\checkmark$} & \textcolor[HTML]{6ebd9b}{$\checkmark$} / \textcolor[HTML]{6ebd9b}{$\checkmark$} & \textcolor[HTML]{6ebd9b}{$\checkmark$} / \textcolor[HTML]{6ebd9b}{$\checkmark$} & \textcolor[HTML]{6ebd9b}{$\checkmark$} / \textcolor[HTML]{6ebd9b}{$\checkmark$} & \textcolor[HTML]{6ebd9b}{$\checkmark$} / \textcolor[HTML]{6ebd9b}{$\checkmark$} & \textcolor[HTML]{9a371e}{$\times$} / \textcolor[HTML]{9a371e}{$\times$} & \textcolor[HTML]{9a371e}{$\times$} / \textcolor[HTML]{9a371e}{$\times$} \\
Strengthening the CO2 emission performance targets for new heavy-duty vehicles & \textcolor[HTML]{6ebd9b}{$\checkmark$} / \textcolor[HTML]{6ebd9b}{$\checkmark$} & \textcolor[HTML]{6ebd9b}{$\checkmark$} / \textcolor[HTML]{6ebd9b}{$\checkmark$} & \textcolor[HTML]{6ebd9b}{$\checkmark$} / \textcolor[HTML]{6ebd9b}{$\checkmark$} & \textcolor[HTML]{6ebd9b}{$\checkmark$} / \textcolor[HTML]{6ebd9b}{$\checkmark$} & \textcolor[HTML]{9a371e}{$\times$} / \textcolor[HTML]{6ebd9b}{$\checkmark$} & \textcolor[HTML]{9a371e}{$\times$} / \textcolor[HTML]{9a371e}{$\times$} & \textcolor[HTML]{9a371e}{$\times$} / \textcolor[HTML]{9a371e}{$\times$} \\
Substantiation and communication of explicit environmental claims (Green Claims Directive) & \textcolor[HTML]{6ebd9b}{$\checkmark$} / \textcolor[HTML]{6ebd9b}{$\checkmark$} & \textcolor[HTML]{6ebd9b}{$\checkmark$} / \textcolor[HTML]{6ebd9b}{$\checkmark$} & \textcolor[HTML]{6ebd9b}{$\checkmark$} / \textcolor[HTML]{6ebd9b}{$\checkmark$} & \textcolor[HTML]{6ebd9b}{$\checkmark$} / \textcolor[HTML]{6ebd9b}{$\checkmark$} & \textcolor[HTML]{6ebd9b}{$\checkmark$} / \textcolor[HTML]{6ebd9b}{$\checkmark$} & \textcolor[HTML]{4663ec}{$\blacksquare$} / \textcolor[HTML]{6ebd9b}{$\checkmark$} & \textcolor[HTML]{9a371e}{$\times$} / \textcolor[HTML]{9a371e}{$\times$} \\
Type-approval of motor vehicles and engines with respect to their emissions and battery durability (Euro 7) & \textcolor[HTML]{9a371e}{$\times$} / \textcolor[HTML]{9a371e}{$\times$} & \textcolor[HTML]{9a371e}{$\times$} / \textcolor[HTML]{6ebd9b}{$\checkmark$} & \textcolor[HTML]{9a371e}{$\times$} / \textcolor[HTML]{9a371e}{$\times$} & \textcolor[HTML]{6ebd9b}{$\checkmark$} / \textcolor[HTML]{6ebd9b}{$\checkmark$} & \textcolor[HTML]{6ebd9b}{$\checkmark$} / \textcolor[HTML]{6ebd9b}{$\checkmark$} & \textcolor[HTML]{7e7e7e}{$NO$} / \textcolor[HTML]{6ebd9b}{$\checkmark$} & \textcolor[HTML]{6ebd9b}{$\checkmark$} / \textcolor[HTML]{6ebd9b}{$\checkmark$} \\
Union certification framework for carbon removals & \textcolor[HTML]{9a371e}{$\times$} / \textcolor[HTML]{6ebd9b}{$\checkmark$} & \textcolor[HTML]{6ebd9b}{$\checkmark$} / \textcolor[HTML]{6ebd9b}{$\checkmark$} & \textcolor[HTML]{6ebd9b}{$\checkmark$} / \textcolor[HTML]{6ebd9b}{$\checkmark$} & \textcolor[HTML]{6ebd9b}{$\checkmark$} / \textcolor[HTML]{6ebd9b}{$\checkmark$} & \textcolor[HTML]{6ebd9b}{$\checkmark$} / \textcolor[HTML]{6ebd9b}{$\checkmark$} & \textcolor[HTML]{9a371e}{$\times$} / \textcolor[HTML]{9a371e}{$\times$} & \textcolor[HTML]{9a371e}{$\times$} / \textcolor[HTML]{9a371e}{$\times$} \\
Union-wide effect of certain driving disqualifications & \textcolor[HTML]{6ebd9b}{$\checkmark$} / \textcolor[HTML]{6ebd9b}{$\checkmark$} & \textcolor[HTML]{6ebd9b}{$\checkmark$} / \textcolor[HTML]{6ebd9b}{$\checkmark$} & \textcolor[HTML]{6ebd9b}{$\checkmark$} / \textcolor[HTML]{6ebd9b}{$\checkmark$} & \textcolor[HTML]{6ebd9b}{$\checkmark$} / \textcolor[HTML]{6ebd9b}{$\checkmark$} & \textcolor[HTML]{9a371e}{$\times$} / \textcolor[HTML]{6ebd9b}{$\checkmark$} & \textcolor[HTML]{9a371e}{$\times$} / \textcolor[HTML]{6ebd9b}{$\checkmark$} & \textcolor[HTML]{9a371e}{$\times$} / \textcolor[HTML]{9a371e}{$\times$} \\
Urban wastewater treatment. Recast & \textcolor[HTML]{6ebd9b}{$\checkmark$} / \textcolor[HTML]{6ebd9b}{$\checkmark$} & \textcolor[HTML]{6ebd9b}{$\checkmark$} / \textcolor[HTML]{6ebd9b}{$\checkmark$} & \textcolor[HTML]{6ebd9b}{$\checkmark$} / \textcolor[HTML]{6ebd9b}{$\checkmark$} & \textcolor[HTML]{6ebd9b}{$\checkmark$} / \textcolor[HTML]{6ebd9b}{$\checkmark$} & \textcolor[HTML]{6ebd9b}{$\checkmark$} / \textcolor[HTML]{6ebd9b}{$\checkmark$} & \textcolor[HTML]{9a371e}{$\times$} / \textcolor[HTML]{6ebd9b}{$\checkmark$} & \textcolor[HTML]{9a371e}{$\times$} / \textcolor[HTML]{9a371e}{$\times$} \\
Wholesale energy market: Union’s protection against market manipulation & \textcolor[HTML]{6ebd9b}{$\checkmark$} / \textcolor[HTML]{6ebd9b}{$\checkmark$} & \textcolor[HTML]{6ebd9b}{$\checkmark$} / \textcolor[HTML]{6ebd9b}{$\checkmark$} & \textcolor[HTML]{6ebd9b}{$\checkmark$} / \textcolor[HTML]{6ebd9b}{$\checkmark$} & \textcolor[HTML]{6ebd9b}{$\checkmark$} / \textcolor[HTML]{6ebd9b}{$\checkmark$} & \textcolor[HTML]{6ebd9b}{$\checkmark$} / \textcolor[HTML]{6ebd9b}{$\checkmark$} & \textcolor[HTML]{6ebd9b}{$\checkmark$} / \textcolor[HTML]{6ebd9b}{$\checkmark$} & \textcolor[HTML]{4663ec}{$\blacksquare$} / \textcolor[HTML]{9a371e}{$\times$} \\
\bottomrule
\end{tabular}
\end{adjustbox}
\caption{\textbf{Comparison of Actual vs. Predicted Group Lines.} We show the actual group lines on the left and the predicted group line by the best performing LLama 3-70B approach on the right for each group. Legend: FOR(\textcolor[HTML]{6ebd9b}{$\checkmark$}), AGAINST(\textcolor[HTML]{9a371e}{$\times$}), ABSTENTION(\textcolor[HTML]{4663ec}{$\blacksquare$}) and majority of the group did not vote(\textcolor[HTML]{7e7e7e}{$NO$}).}
\label{tab:comparison}
\end{table*}

\begin{figure}[t]
    \centering
    \begin{adjustbox}{width={\columnwidth}}
    \includegraphics{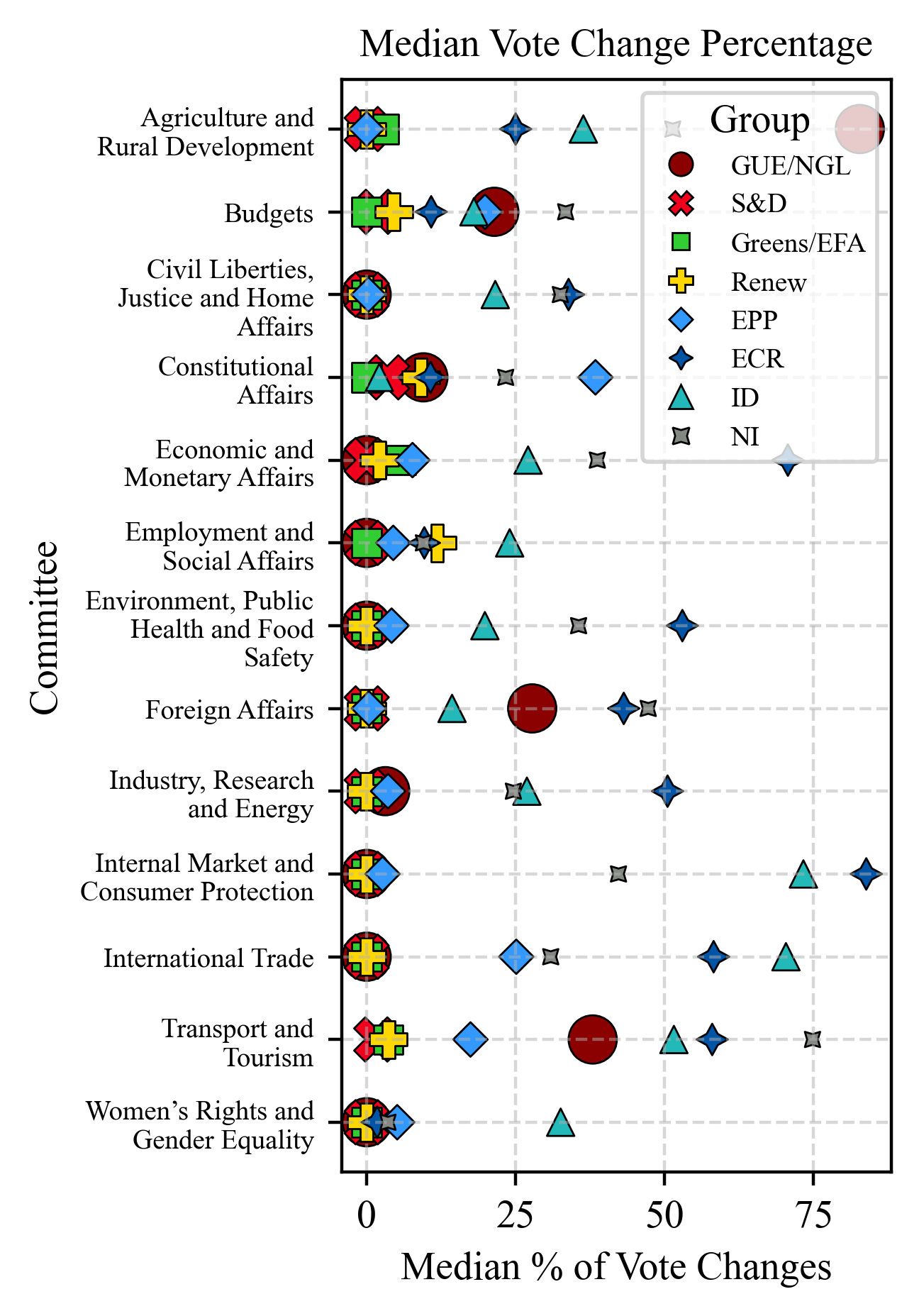} 
    \end{adjustbox}
    \caption{\textbf{Vote changes across groups and policy topics.} We present the median percentage of votes that change when Llama3-70B is prompted with counterfactual speeches. The results are broken down by European groups and categorized by policy topic. While far-left and far-right groups (GUE/NGL, ECR, ID) change their vote significantly more often, the voting behavior of centrist groups (S\&D, Renew, Greens/EFA) remains largely unchanged. This effect seems to be policy specific, as for example models prompted with personas from the far-left group GUE/NGL change their votes more often on Foreign Affairs, but not on proposals concerning Employment and Social Affairs.}
    \label{fig:flip_percentage_70B}
\end{figure}

We analyzed the change in voting behavior across different policy categories when using counterfactual speeches as well. To explore this, we categorize proposals based on the committee that initiated them. Figure \ref{fig:flip_percentage_70B} reveals that individuals from left-leaning (S\&D, Greens/EFA) and even far-left groups (GUE/NGL) consistently maintain their support for proposals on certain topics, such as those from the \textit{Employment and Social Affairs Committee} and the \textit{Women's Rights and Gender Equality Committee}.
For these proposals, most original speeches supported them ($83\%$ and $72\%$, respectively), so the majority of counterfactual debates advise against them. Despite this, the votes of individuals affiliated with left-wing groups remain unchanged. In contrast, center-right (Renew, EPP) and far-right groups (ECR, ID) are more likely to alter their votes on proposals from these committees.

\section{Influence of different attributes}
We first provide details on how we match attributes within the reasoning chains and present additional information and results that are not included in the main section.

\subsection{Attribute Matching}
\label{sec:attribute_matching}

We make use of regular expressions to determine which attributes appear in the reasoning chains of each MEP and to remove specific words from speeches. In the following, we list the substrings used to search for the values of each attribute in the reasoning chains. For each MEP, we only search for their actual attribute values (i.e., we search for \textit{Ireland} and \textit{Irish} if a MEP represents Ireland in the EU parliament). For the attribute \textit{age}, we computed the age of each MEP at the time of voting. We used the exact wording for \textit{birthplace} , which comes from Wikidata\footnote{\href{https://www.wikidata.org/}{https://www.wikidata.org/}}. We also used the exact wording for the politician's \textit{name} and \textit{national party} from the European Parliament Website\footnote{\href{https://www.europarl.europa.eu/meps/en/full-list/all}{https://www.europarl.europa.eu/meps/en/full-list/all}}.

\paragraph{Country}
Luxembourg, Luxembourger, Luxembourgers, Luxembourgian, Luxembourgish, Portugal, Portuguese, Germany, German, Germans, Spain, Spanish, Spaniard, Spaniards, Finland, Finnish, Finn, Finns, Austria, Austrian, Austrians, Belgium, Belgian, Belgians, Netherlands, Dutch, Holland, Italy, Italian, Italians, France, French, Czechia, Czech Republic, Czech, Czechs, Poland, Polish, Hungary, Hungarian, Hungarians, Slovakia, Slovak, Slovaks, Malta, Maltese, Denmark, Danish, Dane, Danes, Slovenia, Slovenian, Slovene, Slovenians, Slovenes, Greece, Greek, Latvia, Latvian, Latvians, Letts, Romania, Romanian, Romanians, Ireland, Irish, Lithuania, Lithuanian, Lithuanians, Bulgaria, Bulgarian, Bulgarians, Croatia, Croatian, Croatians, Croat, Croats, Sweden, Swedish, Swede, Swedes, Cyprus, Cypriots, Cypriot, Estonia, Estonian, Estonians

\paragraph{Gender}
male, Male, man, Man, men, Men, female, Female, woman, Woman, women, Women

\paragraph{Group} 
RENEW, Renew Group, Renew Europe, ALDE, EPP, People's Party, Christian Democrats, Christian Party, Christian Democrat, SD, S\&D, Progressive Alliance of Socialists and Democrats, Social Democrats, Socialists \& Democrats, Socialists and Democrats, Social Democrat, GREEN\_EFA, Greens, Greens/EFA, GREENS/EFA, Green Party, Green Alliance, European Free Alliance, ECR, Reformists, Reformist, GUE\_NGL, GUE/NGL, the Left, The Left, Left group, NI, Non-attached Members, Non-attached Member, ID, Identity and Democracy

\subsection{Additional Attribute Analysis}

Table \ref{tab:attribute_results} presents the weighted F1 scores for different attribute prompt variations. Compared to using only the name, adding the national party and group had the most significant positive impact, achieving performance levels close to those of the full attribute set. Furthermore, country and birthplace contributed modest improvements, while including age and gender slightly reduced performance.

\begin{table}[tb]  
\centering
\begin{adjustbox}{width={\columnwidth}}
\begin{tabular}{lrrrr}
\toprule
 & gender\_mm & gender\_fm & gender\_ff & gender\_mf \\
\midrule
all attributes & 0.10 & 0.06 & 6.20 & 5.74 \\
age & 0.07 & 0.06 & 6.07 & 5.71 \\
birthplace & 0.08 & 0.07 & 6.31 & 6.12 \\
country & 0.05 & 0.03 & 6.69 & 6.21 \\
gender & 0.06 & 0.05 & 6.16 & 5.60 \\
group & 0.04 & 0.02 & 6.23 & 5.69 \\
national party & 0.12 & 0.01 & 6.31 & 5.80 \\
name only & 0.05 & 0.05 & 6.67 & 6.34 \\
\bottomrule
\end{tabular}
\end{adjustbox}
\caption{\textbf{Mentions of gender identities:} We disaggregate the mentions of gender identifiers by gender of the MEPs (e.g. \textit{gender\_fm} indicates the percentage of reasonings by female (\textit{f}) simulated MEPs that mention a male (\textit{m}) identity word). We find that female identity words are mentioned much more often than male identity words. Further, we observe that the reasoning chains contain male and female identity words, respectively, at a similar rate regardless of the gender in the prompt. We conclude that in the reasoning chains, gender identity terms are likely not used to refer to the gender in the prompt (the simulated "own" gender), but rather to discuss a proposal related to gender (e.g. "Combating violence against women and domestic violence").}
\label{tab:gender}
\end{table}

\begin{table}[tb]  
\centering
\begin{tabular}{lr}
\toprule
Group & \% mentions \\
\midrule
GUE/NGL & 23.7 \\
S\&D & 20.7 \\
Greens/EFA & 54.9 \\
Renew & 16.0 \\
EPP & 25.6 \\
ECR & 9.0 \\
ID & 8.62 \\
\bottomrule
\end{tabular}
\caption{\textbf{Percentage of group mentions with only-name prompting:} For only-name-based prompting, we show the fraction of MEPs where the persona prompted LLM mentions their group in the reasoning at least once. The pre-training data seems to contain considerable information about MEPs, without explicit mention in the persona description. We further observe that some groups seem to have a larger share of "known" politicians (e.g. Greens/EFA).}
\label{tab:group_mention_only_name}
\end{table}

\begin{table*}[tb]
\centering
\begin{adjustbox}{width=\linewidth}

\begin{tabular}{lccccccccc}
\toprule
Persona descr. & \textbf{Overall} & GUE/NGL & S\&D & Greens/EFA & Renew & EPP & ECR & ID & NI \\
\midrule
all attributes & \textbf{0.728} & 0.618 & 0.901 & 0.795 & 0.888 & 0.788 & 0.436 & 0.411 & 0.503 \\
\midrule
national party & 0.722 & 0.609 & 0.887 & 0.785 & 0.867 & 0.774 & 0.483 & 0.441 & 0.512 \\
group & 0.718 & 0.615 & 0.900 & 0.793 & 0.896 & 0.805 & 0.417 & 0.302 & 0.483 \\
country & 0.684 & 0.527 & 0.890 & 0.754 & 0.890 & 0.782 & 0.380 & 0.208 & 0.464 \\
birthplace & 0.683 & 0.528 & 0.874 & 0.745 & 0.874 & 0.786 & 0.404 & 0.241 & 0.464 \\
gender & 0.682 & 0.52 & 0.885 & 0.745 & 0.885 & 0.787 & 0.378 & 0.224 & 0.453 \\
\midrule
only name & 0.681 & 0.527 & 0.885 & 0.746 & 0.883 & 0.778 & 0.382 & 0.222 & 0.450 \\
\midrule
age & 0.674 & 0.541 & 0.882 & 0.760 & 0.873 & 0.758 & 0.371 & 0.210 & 0.468 \\
\bottomrule
\end{tabular}
\end{adjustbox}
\caption{\textbf{Prediction performance of voting behavior for prompting with different attributes.} We show the weighted F1 scores for prediction with Llama3-8B. We either prompt with only the name or the name together with the shown attribute(s). We find that using names in conjunction with age performs worst, while using all attributes leads to the best result. National party and European group seem to have the most beneficial effect on the performance. National party improves performance for the far-right groups (ECR, ID) the most.}
\label{tab:attribute_results}
\end{table*}

Table \ref{tab:gender} shows the gender in the reasonings. We disaggregate the mentions of gender identifiers by the gender of the MEPs (e.g., \textit{gender\_fm} indicates the percentage of reasonings by female (\textit{f}) simulated MEPs that mention a male (\textit{m}) identity word). We find that the percentages remain roughly consistent across all prompting strategies and that female identity words are mentioned much more often than male identity words. Further, we observe that the reasoning chains contain male and female identity words at a similar rate, regardless of the gender in the prompt. We conclude that in the reasoning chains, gender identity terms are likely not used to refer to the gender of the persona (the simulated "own" gender), but rather to discuss a proposal related to gender (e.g., "Combating violence against women and domestic violence") or other proposals initiated by the Committee of Women's Rights and Gender Equality.

Table \ref{tab:group_mention_only_name} presents the number of politicians for whom the correct group is mentioned at least once in the reasoning when only their names are provided as personas. The pre-training data appears to contain information on a significant portion of politicians; however, the distribution is biased towards center parties.
Although the pre-training data likely contains more information than what is provided in the prompt, we still observe a significant change in behavior when including all attributes in the prompt instead of just the name. This suggests that using only the name is not always sufficient to guide the model toward a known public figure.

\section{Other models}
\label{sec:other_models}

To prevent knowledge leakage we want to restrict used models to pretraining data cutoff before the date of our earliest vote in our dataset (16. January 2024) to the best of our abilitites and have comparable performance to LLama-3 and Qwen2.5. We identify two other models that fit this criteria: Phi-mini (\texttt{Phi-3.5-mini-instruct} \cite{abdin2024phi} and (\texttt{Mixtral-8x7B-Instruct-v0.1}) \cite{jiang2024mixtral}. However, neither of these models complies with the prompt format nor refuses to answer.

Phi refuses to answer in the required format in the majority of cases. In a test run of 544 * 3 = 1632 inferences, the model refused to comply with the format in 717 of them. In 110 cases, the model outright refuses to answer with the phrase “As an AI language model, …”. Therefore, we do not pursue Phi-mini further.

Out of 27700 votes * 3 runs = 83310 inferences, Mixtral refuses to answer in 86 votes (we prompted multiple times to account for the inconsistency of the output. In these cases, the model does not comply, even when repeatedly prompted). Additionally, it does not comply with the answer restrictions in 275 cases, giving different answers, e.g., “Undecided”. Disregarding such cases, the model achieves a weighted F1-Score of 0.698 when prompted with wikipedia prompts and reasoning, which is comparable to the performance of Llama3-8B with wikipedia prompts and no reasoning. Due to the increased computational costs because of the larger model size, repeated prompting, and worse performance, partly due to the non-compliance with the answer format, we do not conduct further experiments with the model.

\section{Prompting Examples}
\label{sec:prompting_examples}
\begin{figure*}[hbt]
    \centering
    \includegraphics[width=\linewidth]{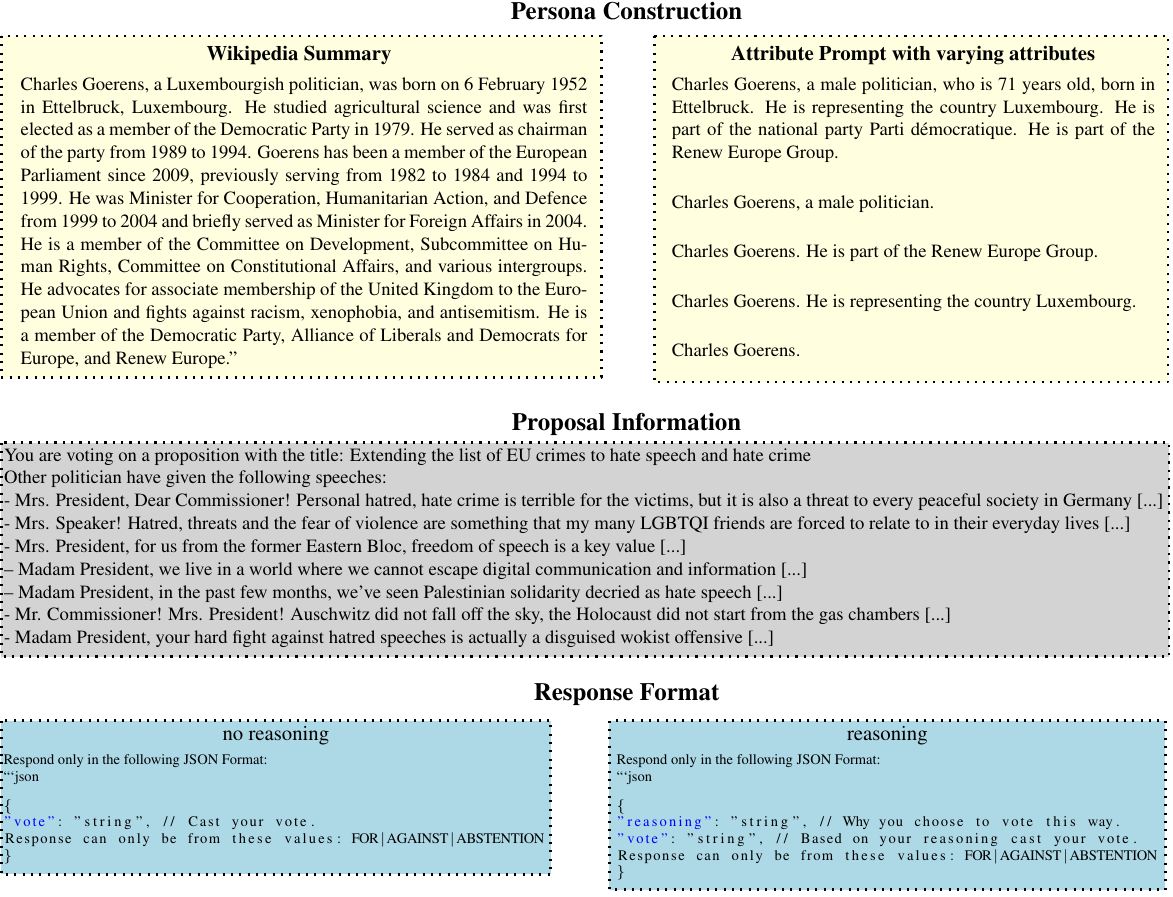}
    \caption{\textbf{Prompting example.} This shows the different approaches we take to prompt the model. We give a persona description, which is either a Wikipedia Summary or an attribute prompt, proposal information in the form of debates (only real speeches are depicted here) and instructions on how to format the response.}
    \label{fig:prompting}
\end{figure*}

We show our different approaches to prompting in Figure \ref{fig:prompting}. We either construct the persona using a summary of the Wikipedia article of the politician generated with Llama3-70B, or we construct the persona with various attributes, consisting of the full name, gender, age, birthplace, country, national party, and group of the politician. Then we provide the proposal information, either with the real speeches of the European Parliament or the counterfactual ones created with Llama3-70B. Then we instruct the model to respond in JSON-Format and either to respond directly with the vote or to first generate a reasoning chain before casting a vote.

\begin{figure*}[t]
    \centering
    \includegraphics[width=\linewidth]{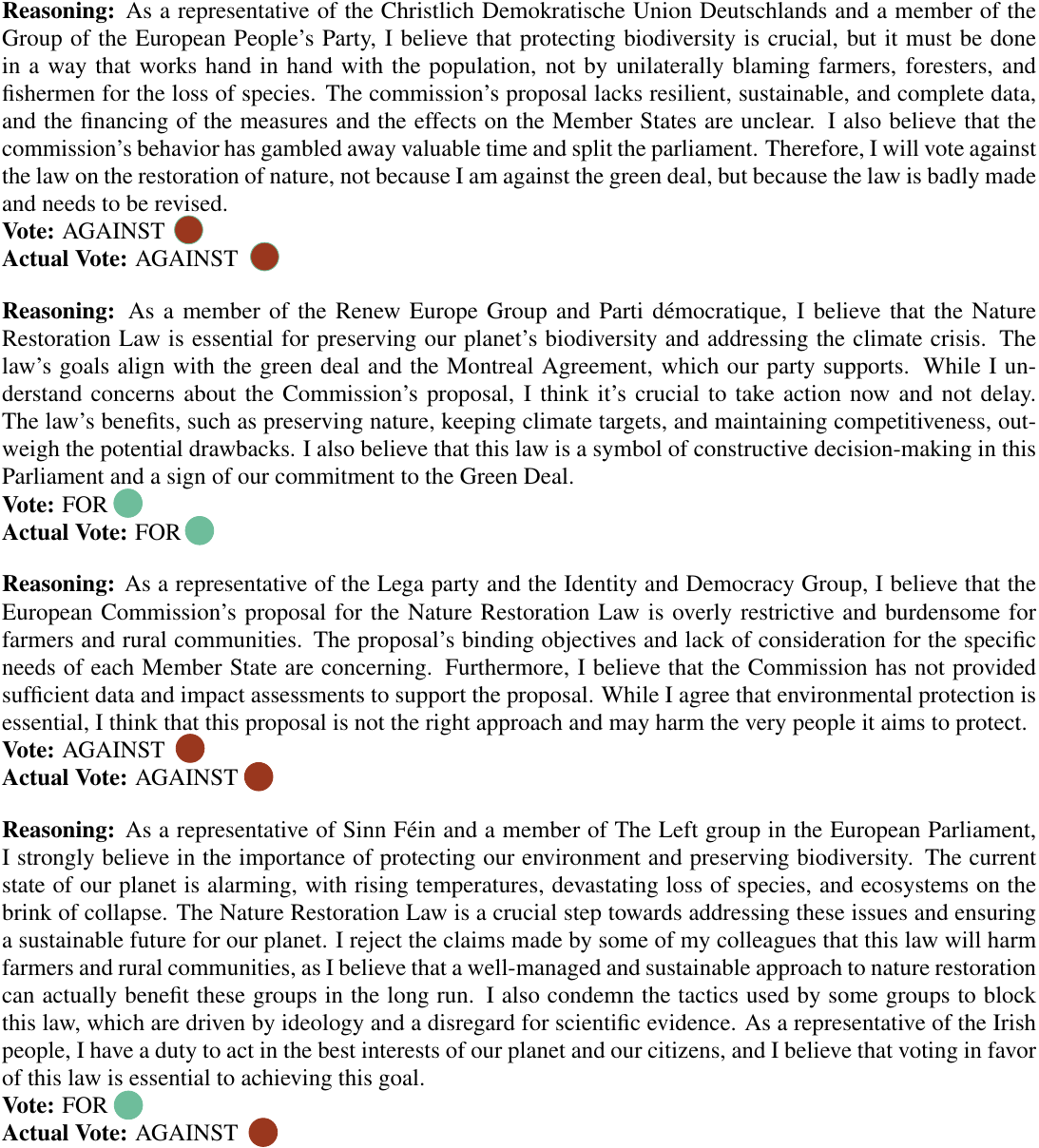}
    \caption{\textbf{Examples of predictions with reasoning.} Four examples of generated predictions with attribute prompts and reasoning by Llama3-70B compared to the ground truth on the Nature Restoration proposal. We see that the models often mention their national party and group and sometimes the nation they represent. Generally the stance of the reasoning chain aligns with the vote, but we have done no qualitative analysis. More information about the roll-call vote can be found here: \href{https://howtheyvote.eu/votes/164499}{https://howtheyvote.eu/votes/164499}.}
    \label{fig:reasoning_examples}
\end{figure*}

Examples of responses are presented in Figure \ref{fig:reasoning_examples}. Models frequently mention the national party and European group associated with their persona, and sometimes reference their nationality.

\begin{figure*}[t]
    \centering
    \includegraphics[width=\linewidth]{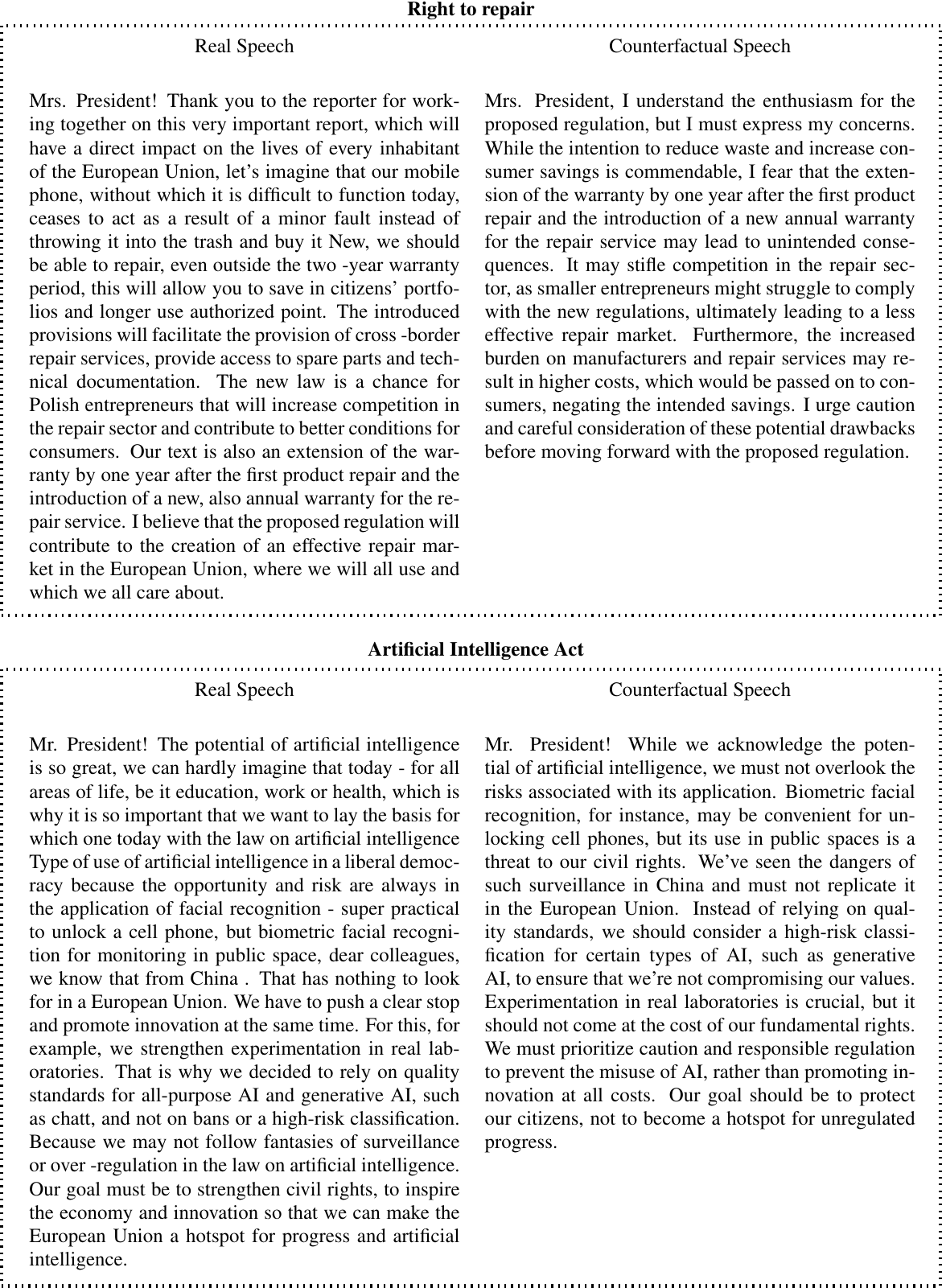}
    \caption{\textbf{Examples of counterfactual speeches.} Two examples of generated counterfactual speeches by Llama3-70B compared to the real speeches they are based on. In both speeches Llama detects the positive stance and creates two stances arguing against it, while staying close to the original arguments.}
    \label{fig:counterfactual_examples}
\end{figure*}

We show two real speeches from the European Parliament and the counterfactual speeches generated by the model in Figure \ref{fig:counterfactual_examples}. We use the prompt \textit{In the following you are given a speech arguing for, against or being neutral towards a proposal in the EU parliament with the name [Title]. First, indicate which stance the given speech takes. Then, generate a new speech that takes the opposite stance (or stays neutral) and only use arguments and information that appear in the given speech.} to determine the stance the LLM has towards the speech and to generate the counterfactual speech. To verify whether the generated counterfactual speeches accurately reflect the opposite stance, three voluntary annotators rate a randomly chosen speech for each of the 47 proposals. We give the voluntary annotators the same first sentence as instruction to determine the stance of the speech. Based on a majority vote, 43 out of 47 speeches indeed align with the opposite stance relative to the original speeches, with a high level of agreement among annotators (Fleiss' Kappa score of $0.800$).

\section{Compute Infrastructure and Time}

We utilize a dual-server setup, each equipped with two \texttt{NVIDIA H100} GPUs (80 GB VRAM) and two \texttt{AMD EPYC 9124 16-Core} processors. On average, a full simulation without reasoning takes approximately 3.5 hours for Llama3-8B and Qwen-7B on a single GPUs, around 18 hours for Llama3-70B and 21 hours for Qwen-72B on two GPUs. However, when reasoning is included, the runtime increases to roughly 17 hours for Llama3-8B and Qwen-7B on a single GPU and approximately 81 hours for Llama3-70B and 95 hours for Qwen-72B on two GPUs. We approximate another 100 dual-GPU hours for creating the counterfactual speeches.

\begin{table*}[p]
    \centering
    \begin{adjustbox}{width={\textwidth}}
    \begin{tabular}{p{8cm}|p{16.5cm}} 
    \toprule
    \textbf{Committee} & \textbf{Proposals} \\
    \midrule
    Agriculture and Rural Development & \textbullet~Geographical Indications for wine, spirit drinks and agricultural products \\
    \midrule
    Budgets & \textbullet~Resolution on ongoing hearings under Article 7(1) TEU regarding Hungary to strengthen the rule of law and its budgetary implications \\
    & \textbullet~Resolution on the situation in Hungary and frozen EU funds \\
    & \textbullet~Establishing the Strategic Technologies for Europe Platform (‘STEP’) \\
    & \textbullet~Establishing the Ukraine Facility \\
    \midrule
    Civil Liberties, Justice and Home Affairs & \textbullet~Combating violence against women and domestic violence \\
    & \textbullet~Preventing and combating trafficking in human beings and protecting its victims \\
    & \textbullet~Single Permit Directive. Recast \\
    & \textbullet~Report on the Commission’s 2023 Rule of Law report \\
    & \textbullet~Extending the list of EU crimes to hate speech and hate crime \\
    & \textbullet~Definition of criminal offences and penalties for the violation of Union restrictive measures \\
    & \textbullet~Situation of fundamental rights in the European Union - annual report 2022 and 2023 \\
    & \textbullet~Prevention of the use of the financial system for the purposes of money laundering or terrorist financing \\
    & \textbullet~European Health Data Space \\
    & \textbullet~Artificial Intelligence Act \\
    \midrule
    Constitutional Affairs & \textbullet~Resolution on ongoing hearings under Article 7(1) TEU regarding Hungary to strengthen the rule of law and its budgetary implications \\
    & \textbullet~Resolution on the situation in Hungary and frozen EU funds \\
    & \textbullet~Deepening EU integration in view of future enlargement \\
    \midrule
    Economic and Monetary Affairs & \textbullet~Prevention of the use of the financial system for the purposes of money laundering or terrorist financing \\
    & \textbullet~Economic governance: requirements for budgetary frameworks of the Member States \\
    & \textbullet~Instant payments in euro \\
    \midrule
    Employment and Social Affairs & \textbullet~Limit values for lead and its inorganic compounds and diisocyanates \\
    \midrule
    Environment, Public Health and Food Safety & \textbullet~European Health Data Space \\
    & \textbullet~Methane emissions reduction in the energy sector \\
    & \textbullet~Packaging and packaging waste \\
    & \textbullet~Fluorinated gases regulation \\
    & \textbullet~Ambient air quality and cleaner air for Europe. Recast \\
    & \textbullet~Urban wastewater treatment. Recast \\
    & \textbullet~Strengthening the CO2 emission performance targets for new heavy-duty vehicles \\
    & \textbullet~Authorisation and supervision of medicinal products for human use and governing rules for the European Medicines Agency \\
    & \textbullet~Union certification framework for carbon removals \\
    & \textbullet~Ecodesign for Sustainable Products Regulation \\
    & \textbullet~Ozone depleting substances \\
    & \textbullet~Plants obtained by certain new genomic techniques and their food and feed \\
    & \textbullet~Type-approval of motor vehicles and engines with respect to their emissions and battery durability (Euro 7) \\
    & \textbullet~Shipments of waste \\
    & \textbullet~Nature restoration \\
    & \textbullet~Substantiation and communication of explicit environmental claims (Green Claims Directive) \\
    \midrule
    Foreign Affairs & \textbullet~Establishing the Ukraine Facility \\
    & \textbullet~Deepening EU integration in view of future enlargement \\
    \midrule
    Industry, Research and Energy & \textbullet~Establishing the Strategic Technologies for Europe Platform (‘STEP’) \\
    & \textbullet~Methane emissions reduction in the energy sector \\
    & \textbullet~Framework of measures for strengthening Europe’s net-zero technology products manufacturing ecosystem (Net Zero Industry Act) \\
    & \textbullet~Energy performance of buildings \\
    & \textbullet~European Digital Identity framework \\
    & \textbullet~Wholesale energy market: Union’s protection against market manipulation \\
    & \textbullet~Cyber Resilience Act \\
    & \textbullet~Energy Charter Treaty:withdrawal of the Union \\
    \midrule
    Internal Market and Consumer Protection & \textbullet~Artificial Intelligence Act \\
    & \textbullet~Substantiation and communication of explicit environmental claims (Green Claims Directive) \\
    & \textbullet~Empowering consumers for the green transition \\
    & \textbullet~Common rules promoting the repair of goods \\
    & \textbullet~Data collection and sharing relating to short-term accommodation rental services \\
    & \textbullet~Prohibiting products made with forced labour on the Union market \\
    \midrule
    International Trade & \textbullet~Energy Charter Treaty:withdrawal of the Union \\
    & \textbullet~Prohibiting products made with forced labour on the Union market \\
    \midrule
    Transport and Tourism & \textbullet~Driving licences \\
    & \textbullet~Union-wide effect of certain driving disqualifications \\
    \midrule
    Women’s Rights and Gender Equality & \textbullet~Combating violence against women and domestic violence \\
    & \textbullet~Preventing and combating trafficking in human beings and protecting its victims \\
    & \textbullet~Inclusion of the right to abortion in the EU Charter of Fundamental Rights \\
    \bottomrule
    \end{tabular}
    \end{adjustbox}
    
    \caption{\textbf{Proposals by Committee:} All proposals in our dataset ordered by the committe that initiated them. We manually assigned the two resolutions and the vote on the "Inclusion of the right to abortion in the EU Charter of Fundamental Rights" to categories. Note that one proposal can be initiated by more than one committee and appear twice in the list.}
    \label{tab:proposals_by_committee}
\end{table*}

\end{document}